%% file: acl_latex.tex
\documentclass[11pt]{article}

\usepackage[final]{acl}

\usepackage{times}
\usepackage{latexsym}

\usepackage[T1]{fontenc}

\usepackage[utf8]{inputenc}

\usepackage{microtype}

\usepackage{inconsolata}

\usepackage{graphicx}

\usepackage{subcaption}

\usepackage{multirow}

\usepackage[table]{xcolor}

\usepackage{todonotes}

\usepackage{cleveref}

\usepackage{tikz}
\usetikzlibrary{shapes.geometric}
\DeclareRobustCommand{\playernode}{%
    \tikz[baseline=-0.65ex]{%
        \node[
            regular polygon, 
            regular polygon sides=3, 
            draw, 
            fill=blue, 
            minimum size=3mm, 
            inner sep=0pt,
            thin
        ] {};
    }%
}

\DeclareRobustCommand{\opponentnode}{
    \tikz[baseline=-1ex]{%
        \node[
            regular polygon,
            regular polygon sides=3,
            draw,
            fill=orange,
            minimum size=3mm,
            inner sep=0pt,
            rotate=180,
            thin
        ] {};
    }%
}

\usepackage{tabularray}

\usepackage{silence}
\title{Communicating Chess Strategies in Natural Language}

\author{
  Langyuan Cui \and Chun Kai Ling \and Hwee Tou Ng\\
  Department of Computer Science\\
  National University of Singapore\\
  13 Computing Drive, Singapore 117417\\
  \texttt{langyuan.c@u.nus.edu}, \texttt{chunkail@nus.edu.sg}, \texttt{dcsnght@nus.edu.sg}\\
  }

\begin{document}
\maketitle
\begin{abstract}
Chess engines have long achieved superhuman playing strength.
However, the underlying strategy behind their move suggestions is difficult for human players, even skilled ones, to comprehend. Motivated by this, we propose the task of chess \textit{strategy verbalization}, which is to describe chess strategies in natural language. We design (i) a pipeline for verbalizing strategies and (ii) an evaluation framework for objective evaluation of generated strategy descriptions. Our experiments show that natural language is a promising and interpretable medium for communicating strategic information to both human and LLM players. We glean additional interesting insights, including (a) the importance of evaluating strategies beyond the main line, (b) the limitations of pure concept-based descriptions, and (c) the limitations of relying on LLMs rather than humans for evaluation.
\end{abstract}

\section{Introduction}

\begin{figure*}[ht]
    \centering
    \includegraphics[width=2\columnwidth]{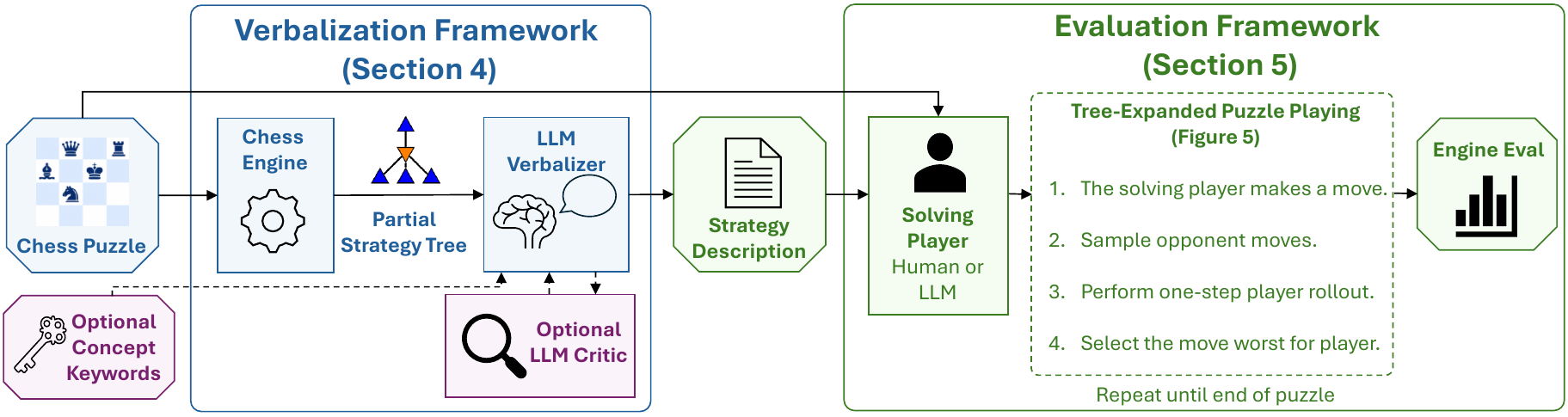}
    \caption{\textbf{Overview of our proposed frameworks.} We selectively extract strategies from chess engines and verbalize them with LLMs, and robustly evaluate descriptions through Tree-Expanded Puzzle Playing (\Cref{fig:eval_tree}) by their downstream utility to players. Each component is discussed in detail in the corresponding sections.}
    \label{fig:framework}
\end{figure*}

Once touted as a bastion of human intellect that would never be matched by machines \cite{hsu1990grandmaster}, chess playing is an activity where machines now dominate. From the classic IBM engine \textit{Deep Blue} \cite{campbell2002deep} to modern bots built upon methods such as \textit{AlphaZero} \cite{silver2017learning}, the sheer strength of chess engines has dramatically transformed the chess landscape.\footnote{After 50 years, the World Computer Chess Championships was retired in 2024 on the grounds that creating stronger engines no longer held any research value \cite{icga2024wccc50}.} 
Yet, while engines excel at move recommendations and evaluations, the crucial question of how to interpret or understand the output of these powerful chess engines is a topic of ongoing research.\footnote{In a recent interview, five-time world champion Viswanathan Anand remarked ``... we [humans] have lost the ability to argue with it [chess engines]. I don't think we have lost the ability to disagree with it, but we have lost the ability to explain why'' \cite{anand2026interview}.}

A wide range of work has studied explaining engine output, including feature learning to explain move recommendations \cite{puri2019explain,fritz2021some,spinnato2025towards}, concept extraction that makes such features more human interpretable \cite{mcgrath2022acquisition,schut2025bridging}, and commentary generation that expresses explanations in fluent natural language \cite{jhamtani2018learning,zang2019automated,kim2025bridging}. 

In this paper, we study chess \textit{strategy verbalization}: the task of faithfully communicating chess engine-derived strategies in natural language.
Our primary focus is to output natural language descriptions of strategies that are pedagogically useful in \textit{amplifying and enhancing the strength of players}. In contrast, existing work in concept and commentary generation seeks to justify why a given move/variation/strategy ``makes sense'', with improvement in quality of play being secondary.

A central challenge here is that existing strategy representations are either too detailed or too abstract. Engines typically output search trees and evaluations that are accurate but cumbersome to interpret. Conversely, high-level concepts are more interpretable but lack detail about concrete moves to be taken --- both now and in the future. Strategy verbalization must therefore balance informativeness and brevity. Furthermore, even evaluating strategy descriptions requires care, as plausible-sounding descriptions can omit key details, leading to highly suboptimal downstream outcomes. 

In our work, we propose frameworks for generating and evaluating chess \textit{strategy verbalization}, as illustrated in \Cref{fig:framework}. Our main contributions are as follows: we (i) introduce chess \textbf{strategy verbalization}, the task of describing engine strategies in natural language, (ii) propose a verbalization framework that combines chess engines and LLMs to selectively describe important branches of the strategy, (iii) develop an evaluation framework that measures how well a player can follow a description under both main-line and off-main-line opponent responses, and (iv) empirically demonstrate that our proposed approach generates useful descriptions for both LLMs and human players.

\section{Related Work}
\textbf{Chess Commentary Generation}
focuses on post-hoc explanations for a single realized move \cite{jhamtani2018learning,zang2019automated,lee2022improving}. In contrast, our strategies describe how a player should act over \textit{possible future states}, involving reasoning about future trajectories. 
Furthermore, existing works utilize metrics such as BLEU \cite{papineni2002bleu}, while recently \citet{kim2025bridging} use LLM-as-a-judge to assess generated commentaries. Rather than evaluating the surface form of descriptions, our framework evaluates descriptions via downstream utility, measuring pedagogical usefulness in a more principled manner. 

\paragraph{Concept Learning.}
This approach extracts concepts from chess engines' representations \cite{mcgrath2022acquisition,schultz2024mastering,palsson2023unveiling}. The concepts involve chess-specific terms such as names of tactics (e.g., forks, skewers, discovered checks) or themes/motifs (e.g., king safety, development, zugzwang). While useful for understanding model behavior, concepts are often too abstract and lack enough detail to communicate an executable strategy.

\paragraph{LLMs for Chess Playing.}
The reasoning capabilities of LLMs have motivated research on whether they can play chess \cite{feng2023chessgpt,schultz2024mastering} with minimal external aid. While LLMs excel in producing plausible-looking moves/explanations, their capacity to explore game trees remains far below that of a modern chess engine \cite{chen2026extracting}, making them substantially weaker players, even when compared to humans.

\paragraph{Evaluating Player Strength.}
Chess evaluation has traditionally emphasized playing strength, often measured by Elo ratings \cite{elo1978rating}. Prior studies of LLM chess-playing ability therefore commonly use head-to-head tournaments and report Elo ratings \cite{zhang2025complete,kolasani2025llm}. In contrast, we work with chess puzzles, which typically do not terminate in a win or loss, but instead use chess engines to estimate the quality of the resulting board states.

\paragraph{Evaluating Strategies.}
Metrics for relative player strength typically do not capture the robustness of a strategy \cite{timbers2022approximate}. Evaluation based on head-to-head games against a fixed set of opponents may obscure failures in parts of the strategy that are not reached during evaluation.
A more robust game-theoretic notion is exploitability, which evaluates a strategy against a worst-case opponent and therefore explores all branches \cite{lanctot2017unified}. Our work mimics this principled stance: we evaluate strategy descriptions beyond the main line, across multiple opponent responses.

\section{What is Chess Strategy Verbalization}
Chess is a two-player, deterministic, alternating turn, perfect-information game over a sequence of board states. A move represents a local decision at a single board state, whereas a \textit{complete} strategy specifies how a player should act at every possible state where it is his turn. Formally, a complete strategy is a mapping from board states to moves. We analogously define a \textit{partial} strategy as a partial mapping. A chess engine implicitly defines a complete strategy that can be constructed by querying for its best move at every state. We refer to this as the \textit{Engine Strategy} (ES). While chess engines are not formally optimal, they are sufficiently accurate that we can treat their move recommendations and 
board state evaluations as the ground truth.

The task of strategy verbalization in chess is to convert an ES into a \textit{natural language} description that enables a player to recover the underlying strategy. This imposes two competing criteria. The description should be (i) \textbf{faithful}, preserving the original strategy so that a player can select the intended move in relevant board states, and (ii) \textbf{compact}, since a strategy is too large to describe exhaustively. Strategy verbalization therefore requires compressing a strategy into language while preserving the strategically important components.

\begin{figure}[t]
    \centering
    \includegraphics[width=\columnwidth]{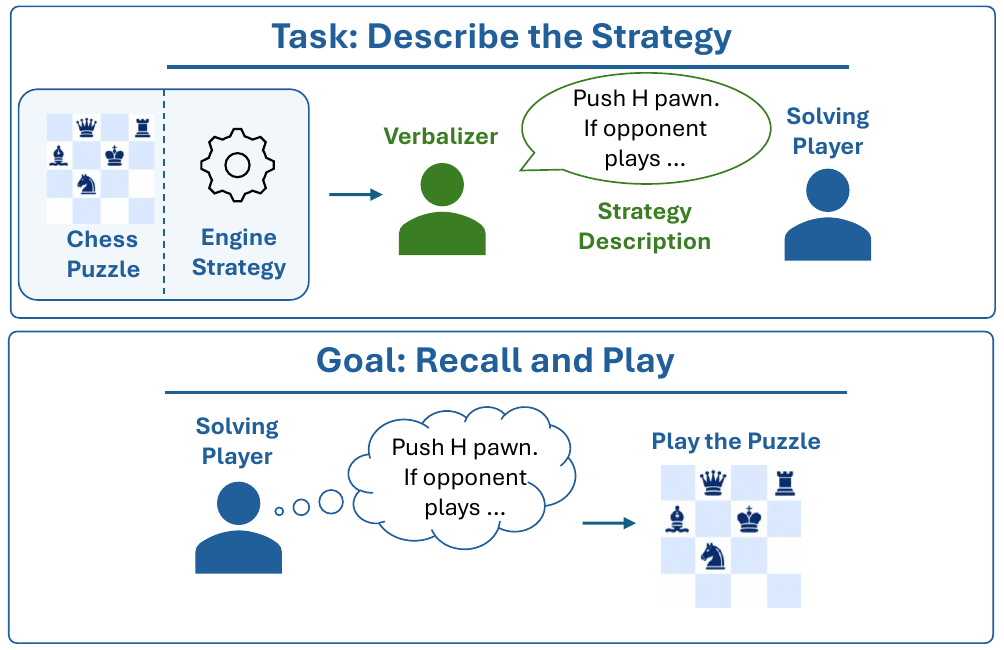}
    \caption{\textbf{Chess strategy verbalization}. The verbalizer outputs a natural language strategy (top) that a player utilizes to perform well in the puzzle (bottom).}
    \label{fig:strat_describe}
\end{figure}

We study strategy verbalization in chess puzzles. 
In puzzles, the quality of the resulting positions varies greatly between optimal and suboptimal play, making them more suitable for experimentation.
Each puzzle is defined by an initial board state and a fixed puzzle length measured in plies, where a ply is a single move by one player. The \textit{solving player} moves first; the opponent moves second. Each puzzle ends with the solving player making the last move. The goal is to describe an ES for the given puzzle for the solving player to make moves in the board states that could be encountered (\Cref{fig:strat_describe}).

\section{A Pipeline for Strategy Verbalization}
Our verbalization framework is organized around two design decisions: how to (i) obtain a partial ES that is relevant to a given puzzle, and (ii) verbalize this selected strategy in natural language so that it remains both faithful and compact enough for downstream use. We address these decisions through the construction of a partial ES tree and a set of LLM-based verbalization methods.

\subsection{Partial Strategy Tree Construction}
\label{sec:strat_tree_construction}
\begin{figure*}[ht]
\begin{minipage}[c]{0.31\textwidth}
\centering
\begin{tikzpicture}[
  level distance=8mm,
  every node/.style={circle,draw,minimum size=2mm,inner sep=0pt},
  level 1/.style={sibling distance=13mm},
  level 2/.style={sibling distance=4mm},
  level 3/.style={sibling distance=6mm},
  invertednode/.style={
    regular polygon,
    regular polygon sides=3,
    draw,
    fill=orange,
    minimum size=3mm,
    inner sep=0pt,
    rotate=180,
    thin,
  },
  uprightnode/.style={
    regular polygon, 
    regular polygon sides=3, 
    draw, 
    fill=blue, 
    minimum size=3mm, 
    inner sep=0pt,
    thin,
  },
  endnode/.style={
    draw=none,
    font=\tiny
  },
]
\usetikzlibrary{trees,shapes.geometric}

\node[uprightnode] {}
  child { node[invertednode] {} 
    child { node[uprightnode] {}
      child { node[endnode] {$0.97$}}
      child { node[endnode] {$0.85$}}
      child { node[endnode] (v1) {$0.67$}}
    }
    child { node[uprightnode] {}}
    child { node[uprightnode] {}}
  }
  child { node[invertednode] {} 
    child { node[uprightnode] {}}
    child { node[uprightnode] {}}
    child { node[uprightnode] {}}
  }
  child { node[invertednode] {} 
    child { node[uprightnode] {}}
    child { node[uprightnode] {}}
    child { node[uprightnode] {}
      child[edge from parent/.style={draw=none}] { node[endnode] (v2) {}}
    }
  };
\path (v1) -- (v2) node [draw=none, black, font=\small, midway, sloped] {$\dots$};

\end{tikzpicture}
\end{minipage}
\begin{minipage}[c]{0.31\textwidth}
\centering
\begin{tikzpicture}[
  level distance=8mm,
  every node/.style={circle,draw,minimum size=2mm,inner sep=0pt},
  level 1/.style={sibling distance=13mm},
  level 2/.style={sibling distance=4mm},
  level 3/.style={sibling distance=6mm},
  invertednode/.style={
    regular polygon,
    regular polygon sides=3,
    draw,
    fill=orange,
    minimum size=3mm,
    inner sep=0pt,
    rotate=180,
    thin,
  },
  uprightnode/.style={
    regular polygon, 
    regular polygon sides=3, 
    draw, 
    fill=blue, 
    minimum size=3mm, 
    inner sep=0pt,
    thin,
  },
  endnode/.style={
    draw=none,
    font=\tiny
  },
]
\usetikzlibrary{trees,shapes.geometric}

\node[uprightnode] {}
  child [edge from parent/.style={draw,thick}]{ node[invertednode] {} 
    child { node[uprightnode] {}
      child [edge from parent/.style={draw,thick}] { node[endnode] {$0.97$}}
      child [edge from parent/.style={draw,thin,dashed}] { node[endnode] {$0.85$}}
      child [edge from parent/.style={draw,thin,dashed}] { node[endnode] (v1) {$0.67$}}
    }
    child { node[uprightnode] {}}
    child { node[uprightnode] {}}
  }
  child [edge from parent/.style={draw,dashed}]{ node[invertednode] {} 
    child { node[uprightnode] {}}
    child { node[uprightnode] {}}
    child { node[uprightnode] {}}
  }
  child [edge from parent/.style={draw,dashed}]{ node[invertednode] {} 
    child { node[uprightnode] {}}
    child { node[uprightnode] {}}
    child { node[uprightnode] {}
      child { node[endnode] (v2) {}
      edge from parent[draw=none]
      }
    }
  };
\path (v1) -- (v2) node [draw=none, black, font=\small, midway, sloped] {$\dots$};

\end{tikzpicture}
\end{minipage}
\begin{minipage}[c]{0.31\textwidth}
\centering
\begin{tikzpicture}[
  level distance=8mm,
  every node/.style={circle,draw,minimum size=2mm,inner sep=0pt},
  level 1/.style={sibling distance=13mm},
  level 2/.style={sibling distance=4mm},
  level 3/.style={sibling distance=6mm},
  invertednode/.style={
    regular polygon,
    regular polygon sides=3,
    draw,
    fill=orange,
    minimum size=3mm,
    inner sep=0pt,
    rotate=180,
    thin,
  },
  uprightnode/.style={
    regular polygon, 
    regular polygon sides=3, 
    draw, 
    fill=blue, 
    minimum size=3mm, 
    inner sep=0pt,
    thin,
  },
  endnode/.style={
    draw=none,
    font=\tiny
  },
]
\usetikzlibrary{trees,shapes.geometric}

\node[uprightnode] {}
  child [edge from parent/.style={draw,thick}]{ node[invertednode] {} 
    child { node[uprightnode] {}
      child [edge from parent/.style={draw,thick}] { node[endnode] {$0.97$}}
      child [edge from parent/.style={draw,thin,dashed}] { node[endnode] {$0.85$}}
      child [edge from parent/.style={draw,thin,dashed}] { node[endnode] (v1) {$0.67$}}
    }
    child { node[uprightnode] {}}
    child [edge from parent/.style={draw,thin,dashed}]{ node[uprightnode] {}}
  }
  child [edge from parent/.style={draw,dashed}]{ node[invertednode] {} 
    child { node[uprightnode] {}}
    child { node[uprightnode] {}}
    child { node[uprightnode] {}}
  }
  child [edge from parent/.style={draw,dashed}]{ node[invertednode] {} 
    child { node[uprightnode] {}}
    child { node[uprightnode] {}}
    child { node[uprightnode] {}
      child { node[endnode] (v2) {}
      edge from parent[draw=none]
      }
    }
  };
\path (v1) -- (v2) node [draw=none, black, font=\small, midway, sloped] {$\dots$};

\end{tikzpicture}
\end{minipage}
\caption{\textbf{Left:} Game tree for a hypothetical 3-ply puzzle rooted at the initial board state. Nodes \playernode \ (resp.\opponentnode ) represent board states where it is the solving player's (resp. opponent's) turn to move, while leaves contain engine evaluation scores. Edges represent moves, ordered from \textit{best to worst} for the player-to-move from left to right. For simplicity, both players are assumed to have three legal moves. \textbf{Center:} Reachable subtree if the solving player follows ES, shown in bold. \textbf{Right:} Pruned partial strategy tree with $k_{strat}=2$, retaining only the top two actions at each opponent state.}
\label{fig:strat_tree}
\end{figure*}
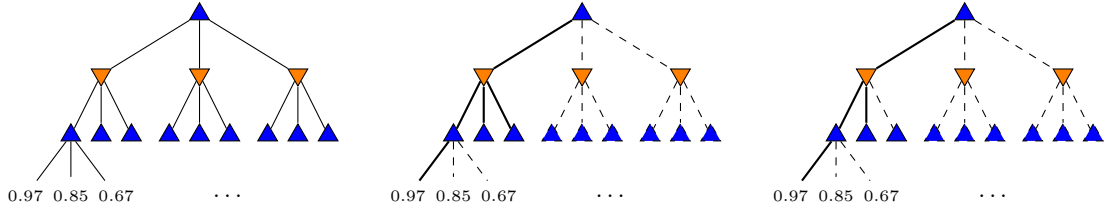

Given a puzzle, we consider the game tree rooted at its initial board state with depth equal to the puzzle length, restricting attention to the reachable states. We further consider a subtree of states reachable when the solving player follows the ES, i.e., playing the best move at each state. Ideally, the partial strategy tree would specify the best move at every solving player node in this subtree. However, as the opponent could choose \textit{any} move, explicitly constructing this subtree is infeasible  
as it grows exponentially in depth, and the opponent's branching factor could be large (typically $\sim 35$ for chess). 
We therefore prune it by retaining at most $k_{strat}$ opponent 
actions at each opponent node; this is given by the top $k_{strat}$ best engine-ranked moves from the opponent's perspective.
The resulting pruned strategy tree (\Cref{fig:strat_tree})
enjoys a far smaller branching factor of $k_{strat}$, striking a balance between compactness and retaining key strategic information in the search tree. Tree construction only requires relatively inexpensive engine calls; thus our approach is practical when $k_{strat}$ is not too large (e.g., $\leq 3$).

\subsection{Verbalizer Design}
\label{sec:verbalizer_design}

We adopt an LLM-based verbalization approach, driven by their ability to generate fluent natural language and prior evidence that they possess sufficient chess knowledge for game reasoning \cite{kolasani2025llm}. We examine three orthogonal design choices: whether the verbalizer is given access to the partial strategy tree, whether it is guided by high-level chess concepts, and whether the generated description is refined through iterative self-reflection. Strategy and concept guidance can be used separately or jointly, while self-reflection can be applied to any generated description.

\paragraph{Strategy Guidance.}
We serialize a partial strategy tree into JSON format and present it alongside the initial board state in Forsyth-Edwards Notation (FEN) to an LLM. The LLM is instructed to describe the strategy tree in the puzzle's context. Exposing the verbalizer to the partial ES allows it to generate a description grounded in a strategy.

\paragraph{Concept Guidance.} \citet{kim2025bridging} studied the use of high-level chess concepts (e.g., forks, king safety) as abstractions to explain moves. 
The focus of \citet{kim2025bridging} was to study how to extract such concepts from a board state. 
To study the role of concepts in isolation, we instead use the keywords that are accurately associated with each puzzle already included in the chess puzzle dataset. These concepts are directly provided to the LLM for strategy description. 

\paragraph{Self-Reflection.}
Inspired by self-reflection \cite{madaan2023self}, we use an LLM critic to evaluate the generated description and provide feedback on whether it adequately conveys the strategy. If the critic judges the description to be lacking, its feedback and the previous generation are passed back to the verbalizer to produce a revised description. This is an independent post-processing step that can be applied to any generated description.  In our experiments, we allow this to repeat for three rounds, using the same LLM as both the verbalizer and critic.
\section{Evaluation Framework}
Prior work used LLM-based evaluation methods such as G-Eval \cite{liu2023g} to assess generated chess commentary along dimensions such as fluency and coherence. However, a plausible-sounding description may still omit important information relevant to the puzzle. We therefore propose a framework that evaluates the faithfulness of a description by its downstream utility (see \Cref{fig:framework}).

\paragraph{Reconstruction-Based Evaluation.}
We view the verbalizer as an encoding strategy into natural language. We employ a \textit{solving player} as an implicit decoder to recover the strategy by playing the puzzle under the guidance of the description, as illustrated in \Cref{fig:framework}. We evaluate each description by its downstream utility, measured using the chess engine's evaluation of the final board state reached by the player. This score is interpreted relative to the player's no-description baseline, allowing us to measure how much the description improves decision making beyond the player's own ability. 
For scalability, we use LLMs as proxies for human players and complement them with human studies.

\begin{figure*}[ht]
    \centering
    \includegraphics[width=2\columnwidth]{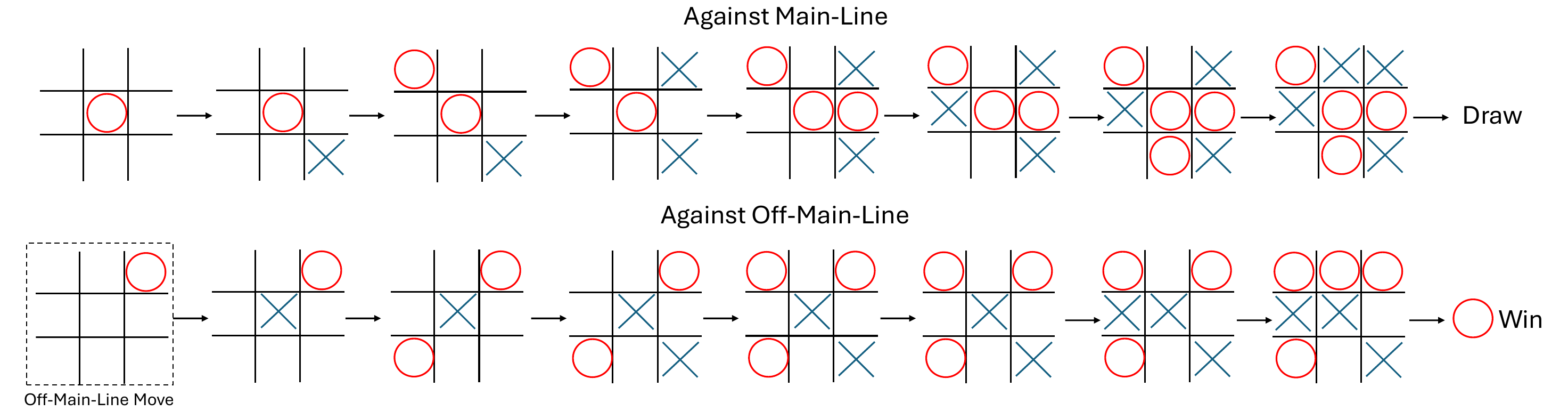}
    \caption{\textbf{Illustration of when a strategy may appear sound against an optimal opponent, but is easily exploitable in off-main-line play.} The cross player is following a strategy that prioritizes moves in the following order: (1) Win if possible, (2) Prevent opponent wins, (3) Center, (4) Corner, (5) Sides, with tiebreaks chosen randomly. This performs well against one strategy (top), but performs poorly against another strategy (bottom).}
    \label{fig:tic_tac_toe}
\end{figure*}

\paragraph{Evaluating Verbalized Strategies.} 
While natural, evaluations based on head-to-head performance against a fixed opponent can be misleading. 
For example, in Rock-Paper-Scissors, playing ``Rock'' obtains a tie against the uniform strategy (which is itself optimal), but loses against an opponent who plays ``Paper''. 
In perfect information games, a strategy may perform well along the main line (e.g., the most common or engine-preferred continuation), while performing poorly against other (possibly suboptimal) \textit{opponent} strategies. 
In \Cref{fig:tic_tac_toe}, we illustrate a strategy for Player 2 in Tic Tac Toe that performs well (i.e., achieves a draw) against the common strategy of starting at the center, but loses to an opponent strategy that starts at the corner. Strategies that perform poorly against \textit{some} opponent strategy are called \textit{exploitable}. From a game-theoretic perspective, exploitability is the gold-standard metric for evaluating strategies \cite{johanson2011accelerating,timbers2022approximate}. 

\begin{figure}
    \centering
    \begin{minipage}[c]{0.11\textwidth}
    \centering
    \begin{tikzpicture}[
      level distance=8mm,
      every node/.style={circle,draw,minimum size=2mm,inner sep=0pt},
      level 1/.style={sibling distance=13mm},
      level 2/.style={sibling distance=4mm},
      level 3/.style={sibling distance=6mm},
      invertednode/.style={
        regular polygon,
        regular polygon sides=3,
        draw,
        fill=orange,
        minimum size=3mm,
        inner sep=0pt,
        rotate=180,
        thin,
      },
      uprightnode/.style={
        regular polygon, 
        regular polygon sides=3, 
        draw, 
        fill=blue, 
        minimum size=3mm, 
        inner sep=0pt,
        thin,
      },
      endnode/.style={
        draw=none,
        font=\tiny
      },
    ]
    \usetikzlibrary{trees,shapes.geometric}
    \node[uprightnode] {}
      child [edge from parent/.style={draw,very thick,red}]{ node[invertednode] {} 
        child [edge from parent/.style={draw=none}]{node[draw=none] {}
          child {node[draw=none,yshift=-4mm] {(1)}}
        }
      };
    \end{tikzpicture}
    \end{minipage}
    \begin{minipage}[c]{0.11\textwidth}
    \centering
    \begin{tikzpicture}[
      level distance=8mm,
      every node/.style={circle,draw,minimum size=2mm,inner sep=0pt},
      level 1/.style={sibling distance=13mm},
      level 2/.style={sibling distance=6mm},
      level 3/.style={sibling distance=6mm},
      invertednode/.style={
        regular polygon,
        regular polygon sides=3,
        draw,
        fill=orange,
        minimum size=3mm,
        inner sep=0pt,
        rotate=180,
        thin,
      },
      uprightnode/.style={
        regular polygon, 
        regular polygon sides=3, 
        draw, 
        fill=blue, 
        minimum size=3mm, 
        inner sep=0pt,
        thin,
      },
      endnode/.style={
        draw=none,
        font=\tiny
      },
    ]
    \usetikzlibrary{trees,shapes.geometric}
    
    \node[uprightnode] {}
      child { node[invertednode] {} 
        child [edge from parent/.style={draw,red}] { node[uprightnode] {}}
        child [edge from parent/.style={draw,red}] { node[uprightnode] {}
          child[edge from parent/.style={draw=none}]{ node[draw=none,yshift=-4mm] {(2)}}
        }
        child [edge from parent/.style={draw,red}] { node[uprightnode] {}}
      };
    \end{tikzpicture}
    \end{minipage}
    \begin{minipage}[c]{0.11\textwidth}
    \centering
    \begin{tikzpicture}[
      level distance=8mm,
      every node/.style={circle,draw,minimum size=2mm,inner sep=0pt},
      level 1/.style={sibling distance=13mm},
      level 2/.style={sibling distance=6mm},
      level 3/.style={sibling distance=6mm},
      invertednode/.style={
        regular polygon,
        regular polygon sides=3,
        draw,
        fill=orange,
        minimum size=3mm,
        inner sep=0pt,
        rotate=180,
        thin,
      },
      uprightnode/.style={
        regular polygon, 
        regular polygon sides=3, 
        draw, 
        fill=blue, 
        minimum size=3mm, 
        inner sep=0pt,
        thin,
      },
      endnode/.style={
        draw=none,
        font=\tiny
      },
    ]
    \usetikzlibrary{trees,shapes.geometric}
    
    \node[uprightnode] {}
      child { node[invertednode] {} 
        child { node[uprightnode] {}
            child [edge from parent/.style={draw,red}] {node[endnode,text=red] {0.75}}
        }
        child { node[uprightnode] {}
            child [edge from parent/.style={draw,red}] {node[endnode,text=red] {0.68}
                child[edge from parent/.style={draw=none}]{ node[draw=none,yshift=4mm] {(3)}}
            }
        }
        child { node[uprightnode] {}
            child [edge from parent/.style={draw,red}]{node[endnode,text=red] {0.21}}
        }
      };
    \end{tikzpicture}
    \end{minipage}
    \begin{minipage}[c]{0.11\textwidth}
    \centering
    \begin{tikzpicture}[
      level distance=8mm,
      every node/.style={circle,draw,minimum size=2mm,inner sep=0pt},
      level 1/.style={sibling distance=13mm},
      level 2/.style={sibling distance=6mm},
      level 3/.style={sibling distance=6mm},
      invertednode/.style={
        regular polygon,
        regular polygon sides=3,
        draw,
        fill=orange,
        minimum size=3mm,
        inner sep=0pt,
        rotate=180,
        thin,
      },
      uprightnode/.style={
        regular polygon, 
        regular polygon sides=3, 
        draw, 
        fill=blue, 
        minimum size=3mm, 
        inner sep=0pt,
        thin,
      },
      endnode/.style={
        draw=none,
        font=\tiny
      },
    ]
    \usetikzlibrary{trees,shapes.geometric}
    
    \node[uprightnode] {}
      child { node[invertednode] {} 
        child [edge from parent/.style={draw,dashed}] { node[uprightnode]  {}
            child [edge from parent/.style={draw, dashed}] {node[endnode] {0.75}}
        }
        child [edge from parent/.style={draw,dashed}] { node[uprightnode] {}
            child [edge from parent/.style={draw, dashed}] {node[endnode] {0.68}
                child[edge from parent/.style={draw=none}]{ node[draw=none,yshift=4mm] {(4)}}
            }
        }
        child [edge from parent/.style={draw,very thick}] { node[uprightnode] {}
            child [edge from parent/.style={draw,very thick}]{node[invertednode] {}}
        }
      };
    \end{tikzpicture}
    \end{minipage}
    \caption{\textbf{Tree-Expanded Puzzle Playing.} (1) The solving player \playernode \ makes a first move. (2) At opponent node \opponentnode, we sample $k_{eval}$ legal moves. (3) For each sampled move, we query the solving player for a response and evaluate the resulting state with a chess engine. (4) We continue with the move that is worst for the solving player. Steps (2)--(4) repeat until the puzzle ends.}
    \label{fig:eval_tree}
\end{figure}
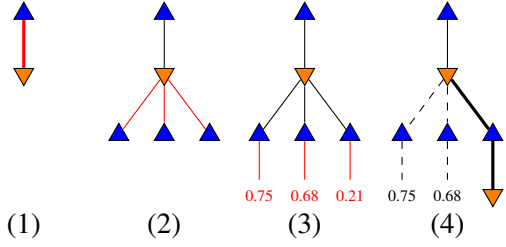

However, measuring exploitability explicitly in our setting is computationally intractable, as it requires constructing the solving player's strategy over all  states reachable from the puzzle's initial state. 
This difficulty is further compounded by the need to query human or LLM players for their moves, which is substantially more expensive than querying a chess engine.
We thus propose \textit{Tree-Expanded Puzzle Playing}, an iterative procedure for evaluating a player’s strategy beyond a single line during puzzle play (\Cref{fig:eval_tree}). At each player node, the solving player makes a move (Step 1). At each opponent node, we consider up to $k_{eval}$ possible actions (Step 2) and roll out each action by one ply to identify the move to which the solving player responds most poorly (Step 3). We then continue expanding only the branch that is worst for the solving player (Step 4), repeating until the end of the puzzle. The final board state identifies the most \textit{exploitable} branch among those explored. Since searching more branches can only lower the corresponding engine evaluation score, the reported value is an upper bound to the player's \textit{worst case} performance.
Importantly, this process remains computationally feasible compared to computing exploitability explicitly since it scales linearly with puzzle length. In our implementation, the $k_{eval}$ opponent moves are chosen as the top moves proposed by the chess engine. While selecting moves based on engine rankings assumes optimal play by the solving player (which may not match the player’s strategy), it serves as a reasonable heuristic for identifying promising branches to explore.

\section{Experiments and Results}
We are primarily interested in answering the following question: \textit{Is the verbalized strategy effective in conveying the engine strategy to a player?} We further examine whether LLMs serve as reasonable proxies for human players, the effect of evaluation beyond the main line, and the benefits of using natural language over other representations for strategy.

\subsection{Experimental Setup}

\paragraph{Models and Chess Engines.}
We use two LLMs for the verbalization module: \texttt{o3} (\texttt{2025-04-16} snapshot), accessed through the OpenAI API \cite{openai_o3_2025}, and \texttt{gpt-oss-120b} \cite{agarwal2025gpt}, hosted locally with vLLM \cite{kwon2023efficient}. We use the same models as solving players in our automatic evaluation. These models were selected based on preliminary experiments showing that they possess sufficient chess knowledge to generate legal moves reliably. In particular, \texttt{o3} has demonstrated chess-playing ability, winning the 2025 Kaggle Game Arena LLM chess tournament \cite{arena2025}. The prompts used are provided in Appendix~\labelcref{app:prompts}. We use the Stockfish 16 \cite{stockfish} chess engine for both strategy construction and evaluation. 

\paragraph{Dataset.}
We evaluate our verbalization framework on the Lichess Puzzle dataset \cite{lichessPuzzles}. Each puzzle consists of a board state in FEN, a puzzle length, and player-voted themes that identify the high-level concepts relevant to the puzzle. We use a subset of 100 puzzles for our experiments. Further details are given in Appendix~\labelcref{app:dataset}.

\paragraph{Baselines.}
When \textit{no verbalizer is used}, we consider the following settings as baselines. 
\textbf{No Strategy}: The player receives only the board state, measuring its baseline playing ability.
\textbf{C}: The player receives the list of concept keywords associated with the puzzle.
\textbf{S}: The player receives the JSON-serialized partial strategy tree.
\textbf{R}: The player receives no strategic information, but is allowed to refine its move at each decision point through self-reflection.
We additionally report an \textbf{Optimal} reference performance, computed by using the chess engine as the solving player and always selecting the engine's best move.

When a \textit{verbalizer is used}, the player is provided with the generated description in \Cref{sec:verbalizer_design}:
\textbf{V-S/V-C/V-S+C}: The verbalizer is given the strategy guidance, concept guidance, or both, respectively.
\textbf{RV-S/RV-C/RV-S+C}: The corresponding descriptions are refined using self-reflection.
We set the branching factor of the constructed strategy tree to $k_{strat}=3$ unless otherwise stated.

\paragraph{Metrics.}
We play the puzzles in each setting using Tree-Expanded Puzzle Playing, setting $k_{eval}=3$ unless otherwise stated. We use Stockfish to evaluate the final board state in terms of centipawn advantage for the solving player, and normalize it into win probability using the Lichess formula \cite{lichessAccuracy}:
$\mathrm{score} = (1 + e^{-0.00368 \cdot cp})^{-1}$
where \(cp\) is the centipawn value. We refer to this value as the \textit{player score}. We additionally report \(\Delta\)Base, the improvement over the \textbf{No Strategy} baseline, and \(\Delta\)Opt, the gap between the player's performance and the \textbf{Optimal} reference. We report the mean and one standard deviation over five runs for each method, with the generated verbalized strategies kept the same across runs.

\subsection{Main Results}
\begin{table*}[ht]
    \centering
    \begin{tabular}{c | c c c | c c c}
    \multirow{2}{*}{\textbf{Method}} & \multicolumn{3}{c|}{Solving Player: \textbf{o3}} & \multicolumn{3}{c}{Solving Player: \textbf{gpt-oss-120b}} \\
        & Player Score & $\Delta$ Base & $\Delta$ Opt & Player Score & $\Delta$ Base & $\Delta$ Opt\\
    \hline
    \multicolumn{7}{c}{\textbf{No Verbalizer}} \\
    \hline
    Optimal & 0.877 (NA) & $+$0.255 & 0 & 0.877 (NA) & $+$0.682 & 0 \\
    No Strategy & 0.622 ($\pm$0.036) & 0 & $-$0.255 & 0.195 ($\pm$0.048) & 0 & $-$0.682\\
    R & \underline{0.698} ($\pm$0.028) & \underline{$+$0.076} & \underline{$-$0.179} &  0.142 ($\pm$0.043) & $-$0.053 & $-$0.735 \\
    C & 0.611 ($\pm$0.041) & $-$0.011 & $-$0.266 & \underline{0.320} ($\pm$0.042) & \underline{$+$0.125} & \underline{$-$0.557} \\
    S & \textbf{0.851} ($\pm$0.009) & \textbf{$+$0.229} & \textbf{$-$0.026} & \textbf{0.602 }($\pm$0.033) & $+$\textbf{0.407} & $-$\textbf{0.275}\\
    \hline
    \multicolumn{7}{c}{\textbf{o3 Verbalizer}} \\
    \hline
    V-C & 0.648 ($\pm$0.040) & $+$0.026 & $-$0.229 & 0.475 ($\pm$0.032) & $+$0.280 & $-$0.402\\
    RV-C & 0.558 ($\pm$0.028) & $-$0.064 & $-$0.319 & 0.472 ($\pm$0.040) & $+$0.277 & $-$0.405\\
    V-S & 0.813 ($\pm$0.037) & $+$0.191 & $-$0.064 &0.580 ($\pm$0.040) & $+$0.385 & $-$0.297\\
    RV-S & \textbf{0.851 }($\pm$0.014) & $+$\textbf{0.229} & $-$\textbf{0.026} & \textbf{0.654} ($\pm$0.055)& $+$\textbf{0.459} & $-$\textbf{0.223} \\
    V-S+C & 0.793 ($\pm$0.028) & $+$0.171 & $-$0.084 & 0.509 ($\pm$0.032) & $+$0.314 & $-$0.368 \\
    RV-S+C & \underline{0.815} ($\pm$0.032) & \underline{$+$0.193} & \underline{$-$0.062} & \underline{0.649} ($\pm$0.047) & \underline{$+$0.454} & \underline{$-$0.228} \\
    \hline
    \multicolumn{7}{c}{\textbf{gpt-oss-120b Verbalizer}} \\
    \hline
    V-C & 0.519 ($\pm$0.040) & $-$0.103 & $-$0.358 & 0.253 ($\pm$0.037) & $+$0.058 & $-$0.624 \\
    RV-C & 0.421 ($\pm$0.043) & $-$0.201 & $-$0.456 & 0.280 ($\pm$0.063) & $+$0.085 & $-$0.597 \\
    V-S & \textbf{0.795 }($\pm$0.041) & $+$\textbf{0.173} & $-$\textbf{0.082} & \underline{0.619} ($\pm$0.044) & \underline{$+$0.424} & \underline{$-$0.258}\\
    RV-S & 0.678 ($\pm$0.044) & $+$0.056 & $-$0.199 & 0.525 ($\pm$0.018) & $+$0.330 & $-$0.352\\
    V-S+C & \underline{0.784} ($\pm$0.041) & \underline{$+$0.162} & \underline{$-$0.093} & \textbf{0.639} ($\pm$0.056) & $+$\textbf{0.444} & $-$\textbf{0.238}\\
    RV-S+C & 0.748 ($\pm$0.038) & $+$0.126 & $-$0.129 & 0.577 ($\pm$0.037) & $+$0.382 & $-$0.300 \\
    \end{tabular}
    \caption{\textbf{Performance of the proposed verbalization methods as evaluated by LLM players.} For each solving player, the best verbalization method is in \textbf{bold} and the second best is \underline{underlined}.}
    \label{tab:chess_main}
\end{table*}

We evaluate the proposed methods on the Lichess Puzzle dataset, with results reported in \Cref{tab:chess_main}.

\paragraph{Effect of Verbalized Strategies.}
Verbalized strategies improve performance over the \textit{No Strategy} baseline, indicating that the generated descriptions convey useful strategic information. This is particularly meaningful given the relatively weak baseline performance, which shows that LLM players remain imperfect without external guidance. However, natural language descriptions typically perform worse than direct access to the JSON strategy representation. This suggests that \textit{verbalization is beneficial but lossy}. Language can compactly summarize the strategy tree, but may omit fine-grained contingencies or describe them in a way that is harder for the solving player to follow.

\paragraph{Effect of Chess Knowledge.}
We observe that (i) chess knowledge of the verbalizer affects the quality of the generated description, while (ii) the playing strength of the solving player affects how it benefits from a description. As a verbalizer, \texttt{o3} generally produces better descriptions that lead to larger improvements, compared to \texttt{gpt-oss-120b}. This difference is less pronounced in the V-S setting, since the strategy tree already specifies the concrete moves. In contrast, when generation is guided only by concepts, the verbalizer's own chess knowledge becomes more important but is much less reliable (see Appendix~\labelcref{app:concept}). Furthermore, the benefit of a description also depends on the solving player's ability, with the weaker \texttt{gpt-oss-120b} often gaining more because there is more room for improvement, while the stronger \texttt{o3} may already produce many correct moves without assistance.

\paragraph{Concept-Guidance and Strategy-Guidance.}
As shown in \Cref{tab:chess_main}, concept-only guidance, whether provided directly or after verbalization, typically yields little improvement. While concepts can serve as hints, \textit{they do not by themselves define an executable strategy}. A notable exception occurs when descriptions in the V-C setting verbalized by \texttt{o3} are provided to \texttt{gpt-oss-120b}, producing a larger performance gain. Inspection of the generated descriptions suggests that this improvement is partly due to \texttt{o3} injecting its own chess knowledge, providing concrete move suggestions rather than merely verbalizing the provided concepts. Examples of generated descriptions are provided in Appendix~\labelcref{app:concept}.

\paragraph{Effect of Self-Reflection.}
Self-reflection has mixed effects. For verbalizers with better chess knowledge, self-reflection can improve descriptions by clarifying ambiguities. For weaker verbalizers, however, self-reflection may degrade performance, often yielding descriptions that are less useful for guiding play. This suggests that \textit{self-reflection is only beneficial when the model can reliably identify and correct strategic omissions}. Examples of descriptions before and after self-reflection are provided in Appendix~\labelcref{app:self_reflection}.

\subsection{Human Evaluation}
\label{sec:human_eval}

\begin{table*}[ht]
    \centering
    \begin{tabular}{c| c c c c c}
    \multirow{2}{*}{\textbf{Method}} & \multicolumn{5}{c}{\textbf{Solving Player}} \\
         & \texttt{o3} & Human & Low Elo & Medium Elo & High Elo \\
    \hline
    No Strategy & 0.387 ($\pm$0.064) & 0.437 ($\pm$0.133) & 0.396 ($\pm$0.100) & 0.457 ($\pm$0.129) & 0.553 ($\pm$0.090)\\
    Human & 0.588 ($\pm$0.047) & \textbf{0.531} ($\pm$0.125) & \textbf{0.494} ($\pm$0.131) & \textbf{0.558} ($\pm$0.120) & \textbf{0.622} ($\pm$0.090)\\
    LLM  & \textbf{0.619} ($\pm$0.043) & 0.513 ($\pm$0.124) & 0.453 ($\pm$0.117) & 0.550 ($\pm$0.082) & 0.581 ($\pm$0.111)\\
    \end{tabular}
    \caption{\textbf{Verbalized strategies evaluated by human players.} $k_{eval} = 3$. The optimal score = 0.722.}
    \label{tab:human_3}
\end{table*}

\begin{table*}[ht]
    \centering
    \begin{tabular}{c| c c c c c}
    \multirow{2}{*}{\textbf{Method}} & \multicolumn{5}{c}{\textbf{Solving Player}} \\
         & \texttt{o3} & Human & Low Elo & Medium Elo & High Elo \\
    \hline
    No Strategy & 0.480 ($\pm$0.057) & 0.452 ($\pm$0.128) & 0.415 ($\pm$0.091) & 0.459 ($\pm$0.128) & 0.569 ($\pm$0.090)\\
    Human  & 0.646 ($\pm$0.037) & 0.573 ($\pm$0.121) & 0.528 ($\pm$0.132) & 0.584 ($\pm$0.114) & \textbf{0.663} ($\pm$0.041)\\
    LLM  & \textbf{0.722} ($\pm$0.000) & \textbf{0.604} ($\pm$0.130) & \textbf{0.556} ($\pm$0.134) & \textbf{0.641} ($\pm$0.091) & 0.657 ($\pm$0.043)\\
    \end{tabular}
    \caption{\textbf{Verbalized strategies evaluated by human players.} $k_{eval} = 1$. The optimal score = 0.722.}
    \label{tab:human_1}
\end{table*}
\begin{figure*}[ht]
    \centering

    \begin{subfigure}[t]{0.49\textwidth}
        \centering
        \includegraphics[width=\linewidth]{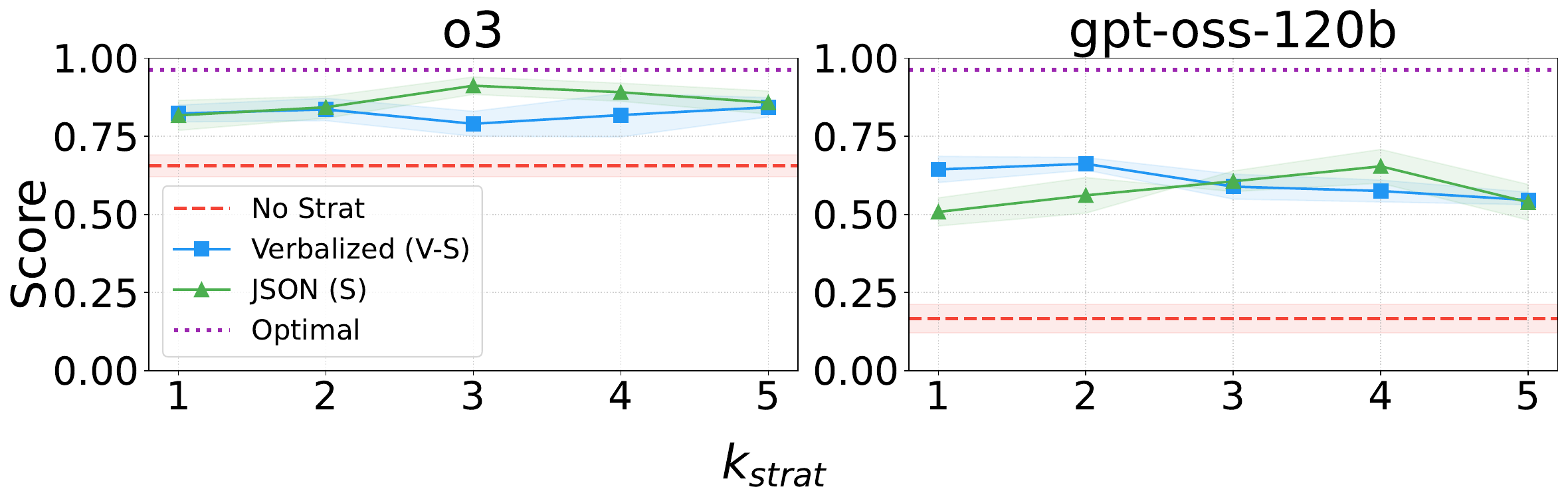}
        \caption{Strategy description verbalized by o3.}
        \label{fig:k_player_o3}
    \end{subfigure}
    \hfill
    \begin{subfigure}[t]{0.49\textwidth}
        \centering
        \includegraphics[width=\linewidth]{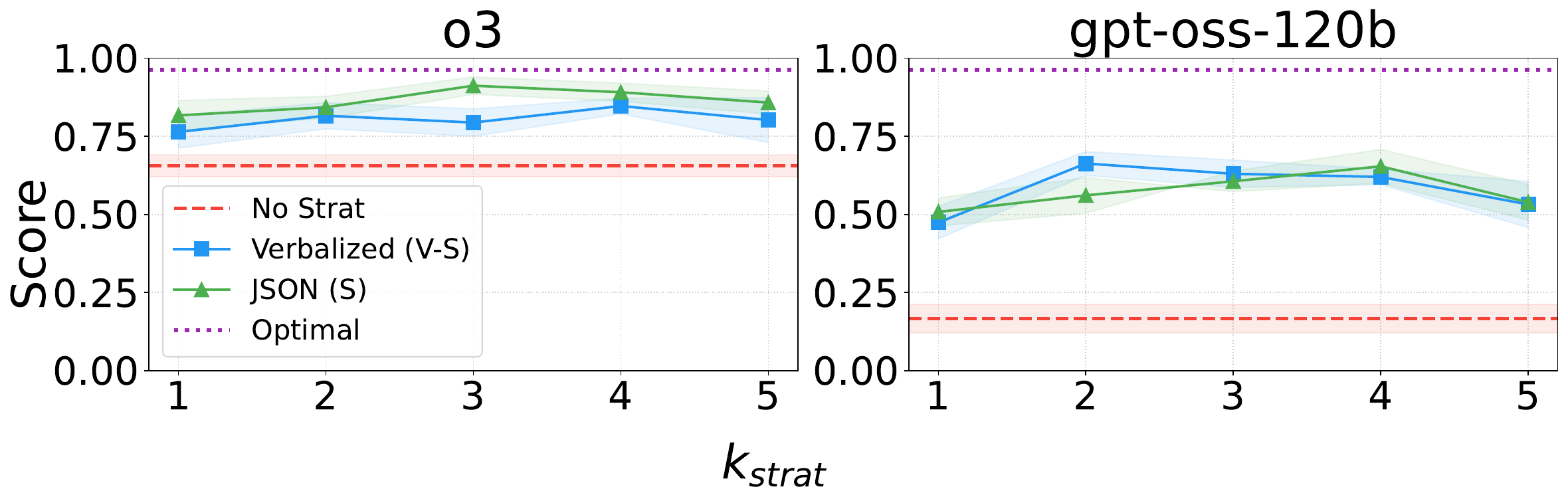}
        \caption{Strategy description verbalized by gpt-oss-120b.}
        \label{fig:k_player_gpt-oss-120b}
    \end{subfigure}

    \caption{\textbf{Performance of solving players as $k_{strat}$ increases}.}
    \label{fig:k_player}
\end{figure*}

\begin{figure}[ht]
    \centering
    \includegraphics[width=\columnwidth]{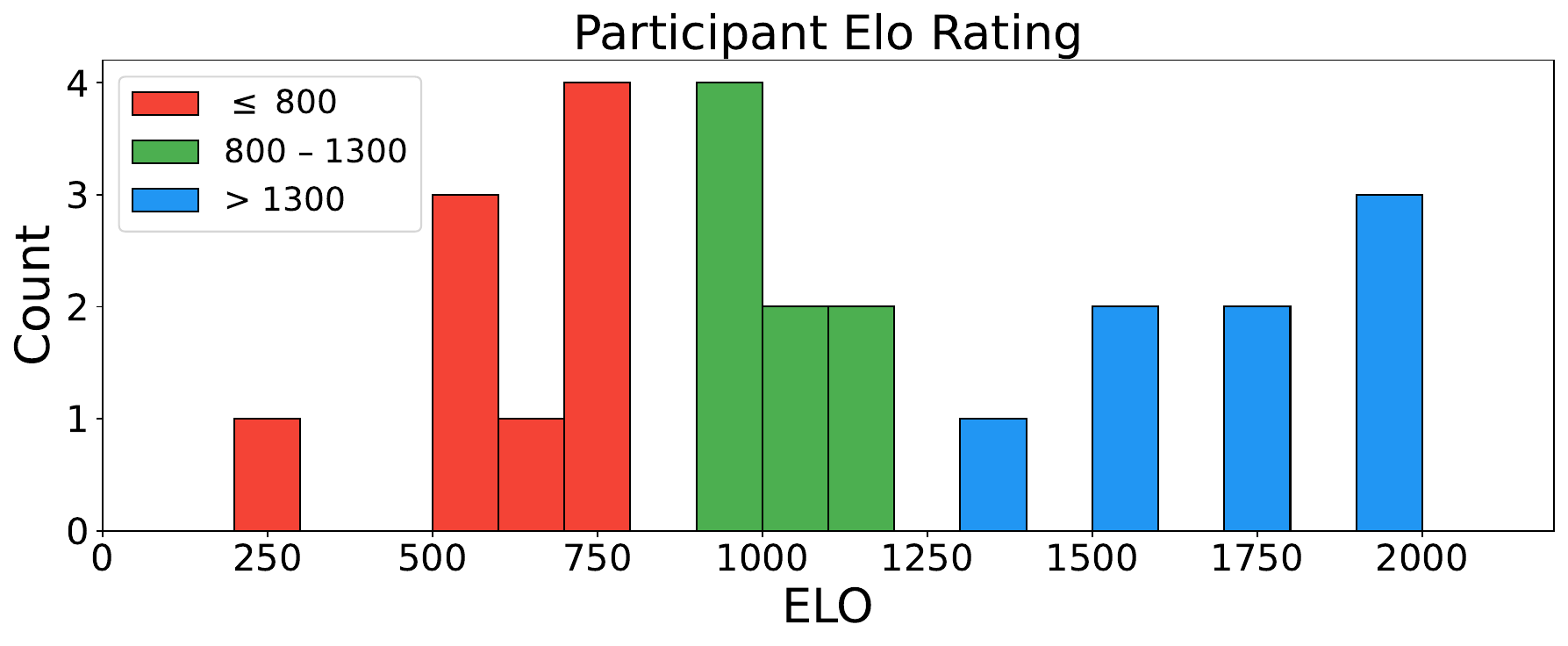}
    \caption{\textbf{Distribution of participants' Chess.com Elo ratings.} Bucket size is set to 100.}
    \label{fig:elo_hist}
\end{figure}

\paragraph{Experimental Setup.} We conducted a human study to examine whether the automatic LLM-based evaluation is aligned with human judgment. We recruited 30 university students to serve as solving players. Among them, 25 participants reported a chess.com \cite{chesscom} Elo rating, which we used to group them into low (\(\leq 800\)), medium (\(801-1300\)), and high (\(1301-2000\)) Elo buckets (see \Cref{fig:elo_hist}). We curated 30 puzzles from a human-written chess manual \citep{chandler2003chess}, together with their solutions, which serve as human-written strategy descriptions. Each human participant is evaluated under three conditions: \textbf{No Strategy}, \textbf{Human-Written Description} (Human), and \textbf{LLM-Generated Description} (LLM), where the description is from the V-S setting by \texttt{o3}. Details are provided in Appendix~\labelcref{app:human_eval}, and results are reported in \Cref{tab:human_3,tab:human_1}. Examples of the strategy descriptions are shown in Appendix~\labelcref{app:human}.

\paragraph{Results Analysis.} Overall, human players benefit from both human-written and LLM-generated descriptions. Interestingly, from \Cref{tab:human_3} we see that human players improve more from human-written than from LLM-generated descriptions, whereas \texttt{o3} improves more from LLM-generated descriptions. This suggests a potential \textit{misalignment between the kinds of descriptions that are useful to humans and to LLMs}. Prior studies have observed similar trends in other settings with LLMs preferring LLM-generated over human-written content \cite{xu2025ai}. In contrast, when \(k_{eval}=1\) in \Cref{tab:human_1}, both human and LLM players benefit more from LLM-generated descriptions. In particular, \texttt{o3} achieves optimal performance with LLM-generated descriptions, suggesting that LLM-generated descriptions are effective in capturing the main line, but may be less successful than human-written ones in communicating off-main-line variations.

\subsection{Ablations and Other Studies}

\begin{figure}[t]
    \centering
    \includegraphics[width=\columnwidth]{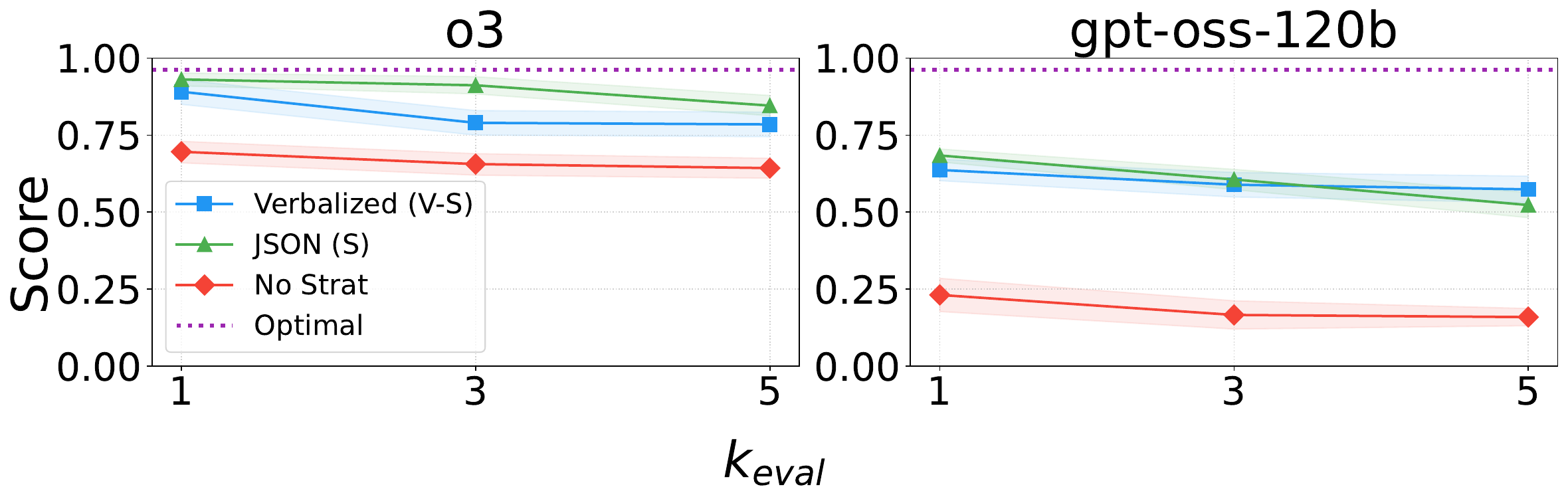}
    \caption{\textbf{Performance as $k_{eval}$ increases}. \textit{V-S} by \texttt{o3}.}
    \label{fig:k_eval}
\end{figure}
\paragraph{Effect of Off-Main-Line Evaluation.}
We vary $k_{eval}$ and report the results in \Cref{fig:k_eval}. Increasing $k_{eval}$ consistently degrades the performance of all methods, illustrating how our proposed evaluation framework exposes flaws in the players' strategy.

\begin{figure}[t]
    \centering
    \includegraphics[width=\columnwidth]{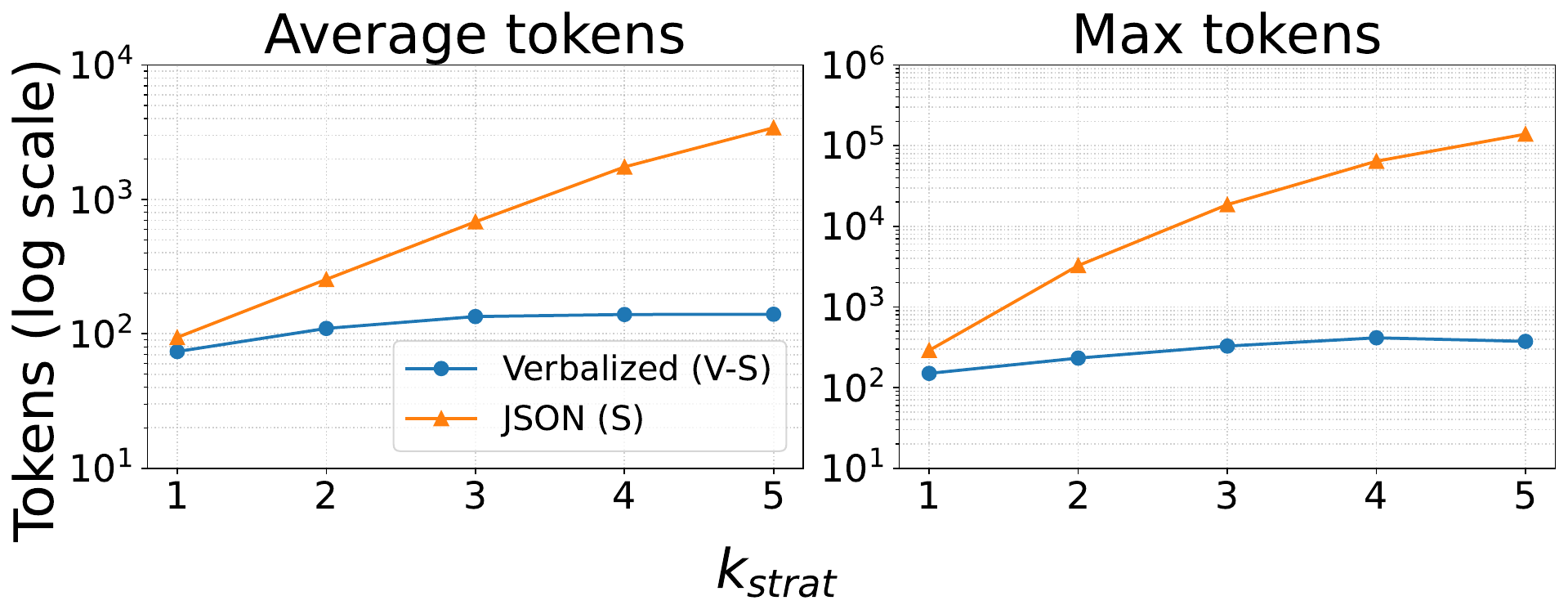}
    \caption{\textbf{Token Count of Verbalized vs JSON Strategy as $k_{strat}$ increases.} Token count measured using \texttt{o3} tokenizer, taken over puzzle instances in one run for each $k_{strat}$.}
    \label{fig:k_strat_tokens}
\end{figure}

\paragraph{Growth in Size of Strategy Representation.}
While JSON effectively conveys strategies as shown in \Cref{tab:chess_main}, its size scales linearly with the strategy tree, which grows exponentially as we include more variations\footnote{In fact, when $k_{strat}=6$, the size of the JSON string for some strategies exceeds the context window of \texttt{o3}.}, as shown in \Cref{fig:k_strat_tokens}. In contrast, the length of the corresponding natural language description grows much slower. Furthermore, \Cref{fig:k_player} shows that the two representations yield generally comparable performance, suggesting that natural language provides a compact yet effective medium for representing strategies. We show the generated descriptions in Appendix~\labelcref{app:tree_size}.

\paragraph{Generalization to Low-Resource Games.} We also study strategy verbalization in Xiangqi, a game with comparatively less online English resources. We discuss the results in Appendix~\labelcref{app:xiangqi}.

\section{Conclusion}
We introduced \textit{chess strategy verbalization}, the task of describing chess strategies in natural language. We proposed several verbalization methods and a framework that robustly evaluates descriptions by their downstream utility. Our results show that natural language descriptions improve both LLM and human play while providing a compact representation of strategy, suggesting that natural language is a promising medium for communicating strategic information. We further highlight the importance of evaluating strategies beyond the main line, and demonstrate the misalignment between LLM-based and human evaluation.
\newpage
\section{Limitations}

\paragraph{The Solving Player as a Confounding Factor.} 
As \Cref{tab:chess_main,tab:human_3} show, the same description can lead to different performance changes depending on the solving player. This is expected, as in practice, the usefulness of a strategy description depends on the recipient. However, it does mean that performance differences between two different verbalizer designs cannot always be attributed solely to the verbalizers themselves. Thus, our framework should be viewed as measuring the effectiveness of a description for a given class of players, rather than as an absolute measure of verbalizer quality.

\paragraph{Misalignment between LLM and Human Preferences.}
As discussed in \Cref{sec:human_eval}, our results suggest that LLMs and humans may benefit from different kinds of strategy descriptions. Our work does not directly address this misalignment. Future work should investigate how to better align LLM-based evaluations with human judgments, as well as how to generate descriptions catered towards human readers.

\paragraph{Evaluating Other Dimensions.}
Our evaluation framework focuses on evaluating how faithfully a description conveys the target strategy. However, there are other factors to consider when evaluating the quality of a description, such as its ease of understanding and naturalness, among others. How strategy descriptions should be evaluated with respect to those dimensions requires further study.

\bibliography{acl_latex}

\appendix

\include{appendix}

\end{document}

%% file: appendix.tex
\section{Additional Related Work}

\paragraph{AI for Recreational Game Playing.}
Research in classic recreational games has resulted in powerful AI systems such as chess engines \cite{campbell2002deep,silver2017learning} and poker bots \cite{brown2018superhuman,brown2019superhuman} that far exceed human performance. With the recent rise of large language models, there has been interest in whether similar progress can be achieved in games involving natural language, such as Werewolf \cite{xu2023exploring}, Diplomacy \cite{meta2022human}, and Avalon \cite{light2023avalonbench}. However, existing AI systems have not yet demonstrated comparable superhuman game-playing performance in these settings. Furthermore, these games are themselves difficult to solve even without the language component \cite{carminati2023hidden}. As such, we focus on studying our task in the more classical domain of chess, where we can rely on chess engines as a proxy for the optimal strategy.

\paragraph{Explainable and Human-Aligned Game Playing.}
As AI systems achieve increasingly strong performance, there is growing interest in making their behavior more understandable and useful to humans. One line of work explains expert or AI-derived policies, including studies on chess commentary generation and our work on strategy verbalization. A second line of work models human play style directly \cite{mcilroy2020aligning,tang2024maia}, highlighting the misalignment between traditional chess AI and human decision-making. A third line of work studies human-AI cooperation, showing that strong agents are not necessarily effective human partners \cite{siu2021evaluation,shih2021critical}. Together, these directions motivate AI systems that are not only strong, but also interpretable, adaptive, and useful to humans.

\paragraph{Describing Structured Artifacts}
Our task is related to work on generating natural language descriptions of structured artifacts. For example, automatic code summarization describes program behavior using specialized models \cite{iyer2016summarizing,hu2018deepcom,feng2020codebert} or agentic LLM systems \cite{luo2024repoagent,yang2025docagent}. Strategies, however, are interactive and adversarial, meaning what matters depends on how the opponent responds. Thus, while these works may provide useful ideas, their direct application is not feasible.

\section{Generalization to Low-Resource Games}

\label{app:xiangqi}
\begin{table}[t]
    \centering
    \begin{tabular}{c| c c}
    \multirow{2}{*}{\textbf{Method}} & \multicolumn{2}{c}{Solving Player: \textbf{o3}} \\
         & Score & $\Delta$ Base \\
    \hline
    \multicolumn{3}{c}{\textbf{No Verbalizer}} \\
    \hline
    Optimal & 0.651 (NA) & $+$0.568 \\
    No Strategy & 0.083 ($\pm$0.027) & $+$0.000\\
    S & 0.453 ($\pm$0.054) & $+$0.370\\
    \hline
    \multicolumn{3}{c}{\textbf{o3 Verbalizer}} \\
    \hline
    V-S & 0.388 ($\pm$0.061) & $+$0.305 \\
    RV-S & 0.341 ($\pm$0.057) & $+$0.258 \\
    \end{tabular}
    \caption{Xiangqi results}
    \label{tab:xiangqi_main}
\end{table}

In our main experiments, we focus on chess because LLMs have demonstrated a reasonable ability to play the game. We additionally examine whether strategy verbalization can be studied in a low-resource game, Xiangqi, also known as Chinese chess. Compared with chess, Xiangqi has substantially fewer English-language resources available online, which likely reduces the amount of relevant training data available to LLMs and limits their ability to play Xiangqi. We use Fairy-Stockfish \cite{fairystockfish} as the Xiangqi engine and use a subset of 50 board states sampled from the test split of the dataset used in \citet{chen2025xiangqi}. Since the dataset does not provide puzzle lengths, we set the number of plies to play to 5 for all puzzles. Results are shown in \Cref{tab:xiangqi_main}.

Despite achieving near-optimal performance when given the JSON strategy in the \textbf{S} setting for chess, the \texttt{o3} solving player shows a much larger performance gap below Optimal in the same setting for Xiangqi. This is interesting because, if the LLM were simply following the JSON strategy as a set of explicit instructions, its performance should remain close to optimal, aside from occasional hallucinations or parsing errors. The larger gap suggests that the model may not be executing the prescribed strategy directly, but instead using it as a hint while relying on its own ability to play. Apart from this difference, \texttt{o3} generally behaves like a weaker solving player in Xiangqi, similar to how \texttt{gpt-oss-120b} behaves in the chess setting.

\section{Qualitative Examples}
\begin{figure}[t]
    \centering
    \includegraphics[width=\columnwidth,page=1]{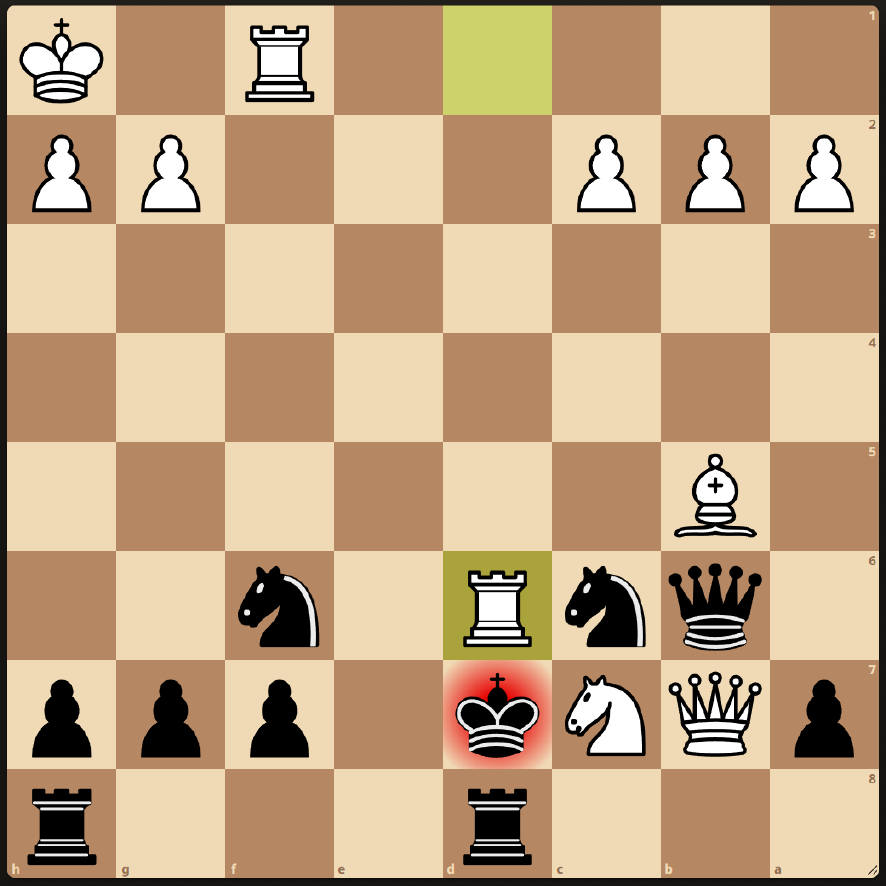}
    \caption{\textbf{A chess puzzle from the Lichess Puzzle Dataset.} Black to move. }
    \label{fig:qualitative}
\end{figure}

In this section, we provide examples of generated strategy descriptions under different settings for the puzzle shown in \Cref{fig:qualitative}.

\subsection{Concept-Guided vs. Strategy-Guided Descriptions}
\label{app:concept}
\begin{table*}[t]
\centering
\begin{tabular}{ c | p{14cm}}

\multicolumn{2}{c}{\textbf{No Verbalizer}} \\
\hline
\textbf{S} & Black: Kxd6, White: Qxb6, Black: axb6 \\
\hline
\textbf{C} & advantage, hangingPiece, middlegame, short\\
\hline
\multicolumn{2}{c}{\textbf{o3 Verbalizer}} \\
\hline
\textbf{V-S} &
Centralize the king with 1. Kxd6, safely capturing the bothersome knight and consolidating the centre. If White tries to counter with 2. Qxb6, calmly reply 2. axb6: the a-pawn recaptures, their queen disappears, and you come out a full piece ahead with a solid position.\\
\hline
\textbf{V-C} &
It's the middlegame: grab the free bishop. Your queen on b6 simply plays Qxb5+, removing White's hanging piece and pocketing material. After the capture consolidate; rooks on the open d- and h-files will safeguard your king and convert the extra piece. Quick, clean advantage.\\
\hline
\textbf{V-S+C} & 
Middlegame clean-up: the rook on d6 is a hanging piece. March the king up with Kxd6 and pocket it. If White answers with Qxb6, just reply axb6; queens come off the board and you emerge a full rook ahead, securing the advantage.\\
\hline
\multicolumn{2}{c}{\textbf{gpt-oss-120b Verbalizer}} \\
\hline
\textbf{V-S} &
Black starts by capturing the white knight on d6 with the king (Kxd6). If White then takes the rook on b6 with the queen (Qxb6), Black simply recaptures the queen with the a-pawn (axb6).
\\
\hline
\textbf{V-C} &
Black is already ahead; White's queen on b7 and bishop on b5 are both hanging. In the middlegame, seize the immediate material gain by playing ...Qxb7 (or ...Rxb5 if more forcing) and then consolidate the extra piece while keeping the king safe. Your aim is a quick win by converting the material advantage, not a long maneuvering game.
\\
\hline
\textbf{V-S+C} & 
Black should immediately capture on d6 with the king to win the hanging piece and secure a material advantage. If White answers by taking the queen on b6, Black simply recaptures with the a-pawn (axb6), restoring material balance and emerging with a winning position in the middlegame."
\\
\end{tabular}
\caption{\textbf{Verbalized Strategies for the V-S/V-C/V-S+C settings.} Puzzle shown in \Cref{fig:qualitative}. $k_{strat}=1$, so only the main line is included. The strategy in the \textbf{S} setting is a condensed version of the actual JSON representation .}
\label{tab:qualitative_concept}
\end{table*}

Verbalized strategies under the V-S/V-C/V-S+C settings are shown in \Cref{tab:qualitative_concept}. When only concept keywords are provided (V-C), the verbalizer must rely on its own ability to infer the concrete moves to recommend. In this case, both \texttt{o3} and \texttt{gpt-oss-120b} propose incorrect moves, suggesting that the black queen should capture the piece on b5 or b7. As shown in \Cref{fig:qualitative}, however, the only legal capture for Black is for the king to capture the rook, i.e., Kxd6. In the V-S+C setting, the LLMs instead use the concept keywords to explain the concrete moves specified by the strategy. Although these explanations are generally coherent, they do not always make the description more compact as intended. Rather, the concepts are often appended to the description instead of being used to summarize it. This highlights an interesting direction for future work: how to use high-level concepts to more effectively compress and communicate the key ideas of a strategy.

\subsection{Effect of Self-Refinement}
\label{app:self_reflection}

\begin{table*}[t]
\centering
\begin{tabular}{ c | p{14cm}}
\hline
\multicolumn{2}{c}{\textbf{o3 Verbalizer}} \\
\hline
\textbf{V-S} &
Start by 1.Kxd6: the king calmly captures the d6-knight, removing the main attacker and coming to safety. If White replies 2.Qxb6+ you accept the queen trade with 2.axb6 and head into a simple ending a rook up. If 2.Bxc6+ you counter with 2.Qxb7, again exchanging queens and remaining a rook ahead. If 2.Ne8+ you meet it with 2.Rhxe8, eliminating the knight and snuffing out all threats. In every line Black has neutralised the attack and emerges with overwhelming extra material.
\\
\hline
\textbf{RV-S} &
1.Kxd6! First, walk the king from d7 to d6 and capture the white rook; you're already winning material and placing the king safely behind your own pieces. If White replies 2.Qxb6, calmly recapture with 2.axb6: queens come off and you stay a full rook ahead.  If 2.Bxc6, answer 2.Qxb7: your queen takes White's queen and you're still up a rook after the minor-piece trade. If 2.Ne8+ (a desperado check), shut it down with 2. Rhxe8: the g8-rook slides left to e8, captures the knight and the checks are over. In every branch you're a rook to the good and heading for an easy endgame win.
\\
\hline
\multicolumn{2}{c}{\textbf{gpt-oss-120b Verbalizer}} \\
\hline
\textbf{V-S} &
Black begins by playing Kxd6, capturing the white knight on d6 with the king. After this capture White has three plausible continuations: 1). White trades queens with Qxb6, and Black recaptures the queen pawn with axb6, simplifying the position. 2).White captures the knight on c6 with Bxc6, after which Black wins the queen on b7 with Qxb7, regaining material.3) White checks with Ne8+, and Black meets the check by exchanging rooks on e8 with Rhxe8, removing the checking piece. In each line Black emerges with a solid material advantage or a simplified, favorable endgame.
\\
\hline
\textbf{RV-S} &
Black is in check from the white rook on d6. The only legal way to meet the check is to capture the rook with the king: \textbf{1.Kxd6}. All other candidate moves are illegal:- King moves to d8, e8, c8, c7, e7, etc. are either occupied or are attacked by white pieces (e.g., e8 is attacked by the white knight on c7, c8 by the white queen on b7, c7 by the same queen, and d8 is occupied by Black's own rook).- No Black piece can interpose on the line d6-d7 because the line consists of only the checking square d6; the squares between the attacker and the king are none. - Capturing the rook with any piece other than the king is impossible: the queen on b6 is blocked by the knight on c6, the rook on d8 is blocked by the king, and neither knight attacks d6.

\textbf{Forced continuation}1.Kxd6  (Black removes the checking piece.)After the capture the white queen on b7 attacks the newly placed Black king on d6 along the diagonal b7-c6-d5-e4 but the knight on c6 blocks that line, so the king is safe for the moment. White's most natural reply is to continue the attack with 2. Qb8+ forcing the king to move again: 2.Qb8+ Kc5  (the only safe square; d5 and e5 are controlled by the white bishop on b5 and the queen on b8 respectively)3.Qb5+ Kd6  (the king returns to the centre) From here Black is materially down (the white rook is gone but White retains a queen, bishop and knight versus Black's queen, two rooks and two knights) and the king is exposed in the centre. There is no viable counter-play that can salvage the position.

\textbf{Conclusion and recommendation} The position is already lost for Black; the only legal move is the forced capture \textbf{1.Kxd6}. After that, White's queen will keep checking and Black's king will be forced to wander in the centre with no compensation. The best practical advice is to play the forced capture and then try to survive the ensuing queen checks, accepting that the game is essentially lost.
\\
\end{tabular}
\caption{\textbf{Verbalized Strategies for the (R)V-S settings.} Puzzle shown in \Cref{fig:qualitative}. $k_{strat}=3$. The description shown for RV-S is after three rounds of refining. Text in the description is put in bold font by the verbalizer itself.}
\label{tab:qualitative_refine}
\end{table*}

Verbalized strategies under the RV-S setting are shown in \Cref{tab:qualitative_refine}. In general, post-refinement descriptions tend to capture more detail, both important and irrelevant, making the descriptions much longer than their pre-refinement versions. This is the most obvious when using \texttt{gpt-oss-120b} as the verbalizer, which produces a much more detailed, but often incorrect, analysis. This ends up causing the strategy to be harder to work with for the solving player.

\subsection{Effect of Strategy Tree Size}
\label{app:tree_size}

\begin{table*}[t]
\centering
\begin{tabular}{ c | p{12.5cm}}
\hline
\multicolumn{2}{c}{\textbf{o3 Verbalizer}} \\
\hline
\textbf{V-S ($k_{strat}=1$)} &
Centralize the king with 1. Kxd6, safely capturing the bothersome knight and consolidating the centre. If White tries to counter with 2. Qxb6, calmly reply 2. axb6: the a-pawn recaptures, their queen disappears, and you come out a full piece ahead with a solid position.
\\
\hline
\textbf{V-S ($k_{strat}=3$)} &
Start by 1.Kxd6: the king calmly captures the d6-knight, removing the main attacker and coming to safety. If White replies 2.Qxb6+ you accept the queen trade with 2.axb6 and head into a simple ending a rook up. If 2.Bxc6+ you counter with 2.Qxb7, again exchanging queens and remaining a rook ahead. If 2.Ne8+ you meet it with 2.Rhxe8, eliminating the knight and snuffing out all threats. In every line Black has neutralised the attack and emerges with overwhelming extra material.
\\
\hline
\textbf{V-S ($k_{strat}=5$)} &
Play 1.Kxd6: the king captures the aggressive rook, untangles your back rank and emerges in the middle surrounded by your own pieces. From here every white try is harmless: 2.Qxb6? - trade queens with 2.axb6 and you stay a whole piece up. 2.Bxc6 - hit back immediately with 2. Qxb7, winning White's queen. 2.Ne8+ - neutralise the check by 2.Rhxe8 and you still have the extra material. 2.Rxf6+ - recapture 2.gxf6; pieces remain even but you still have the spare rook from move one. 2.Qxc6+: return the favour with 2. Qxc6 and simplify. In every branch you keep the material advantage and head for an easily winning end-game."
\\
\end{tabular}
\caption{\textbf{Verbalized Strategies for various values of $k_{strat}$.} Puzzle shown in \Cref{fig:qualitative}.}
\label{tab:qualitative_strat}
\end{table*}

Verbalized strategies for different values of $k_{strat}$ are shown in \Cref{tab:qualitative_strat}. Interestingly, when increasing $k_{strat}$ to 5, the length of the verbalized strategy only roughly doubles, even though the amount of information in the partial strategy tree increases by nearly a factor of four for this puzzle: since it is a 3-ply puzzle, the number of moves in the partial strategy tree increases from 3 to 11. The LLM verbalizer seems to become more concise as the strategy tree grows, cutting out less important narration. For longer puzzles, we also observe that LLMs sometimes omit certain branches of the strategy tree from their descriptions.

\subsection{Human-Written vs. LLM-Generated Descriptions}
\label{app:human}

\begin{figure}[t]
    \centering
    \includegraphics[width=\columnwidth,page=2]{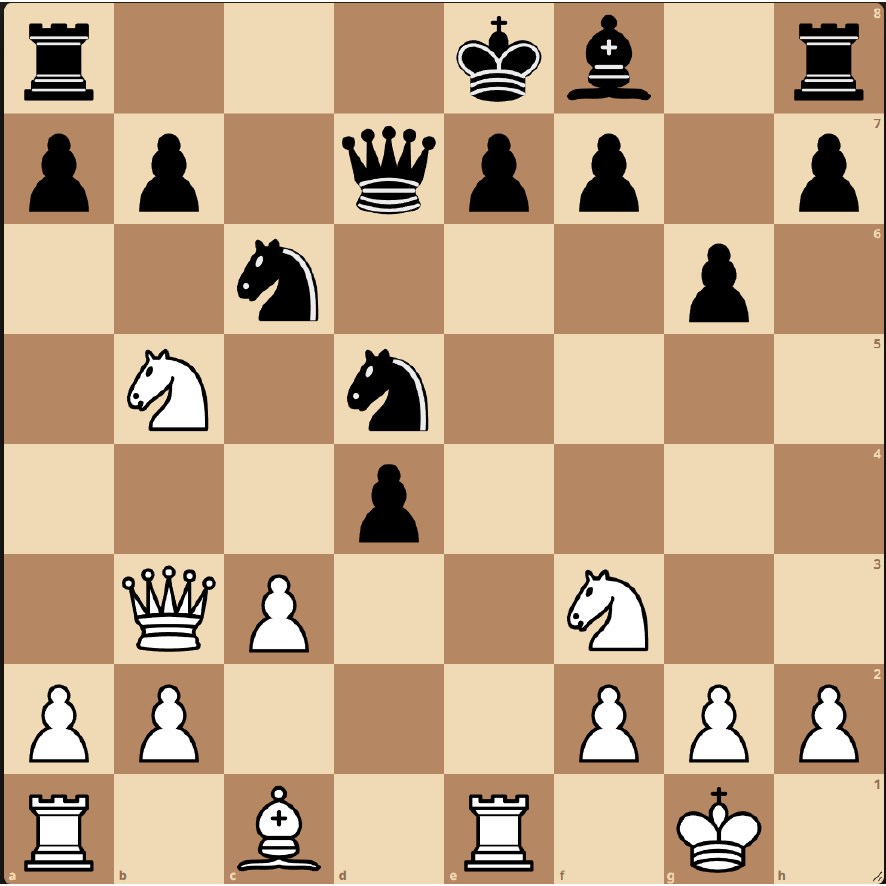}
    \caption{\textbf{A chess puzzle used in human evaluation.} White to move.}
    \label{fig:qualitative_human}
\end{figure}

\begin{table*}[t]
\centering
\begin{tabular}{ c | p{12.5cm}}
\hline
\textbf{Human} &
The temporary sacrifice 1 Qxd5 wins a piece, as after 1...Qxd5 2 Nc7+ White regains the queen with a knight fork.\\
\hline
\textbf{LLM} &
Start by playing 1.Qxd5, grabbing the d-pawn and challenging Black's queen. Three main replies exist: 2 If 1.Qxd5, answer 2.Nc7+, a knight fork that checks the king and will pick up Black's queen next move. If 1.e6 or 1.Rd8 (both defend the queen instead of trading), simply continue 2.Qxd7+ and come out a full queen ahead. In every line White wins decisive material, so 1.Qxd5 is the key move to remember.
\\
\end{tabular}
\caption{\textbf{Strategies shown to participants in human evaluation.} Puzzle shown in \Cref{fig:qualitative_human}. Strategy for LLM is generated by o3 in the V-S setting with $k_{strat}=3$.}
\label{tab:qualitative_human}
\end{table*}

We compare human-written and LLM-generated strategy descriptions in \Cref{tab:qualitative_human}, using the puzzle shown in \Cref{fig:qualitative_human}. Human-written strategies are typically much more concise, focusing only on the key ideas needed to solve the puzzle, whereas LLM-generated descriptions tend to be more verbose and describe every concrete move in detail. LLM-generated descriptions can also contain hallucinations, as noted by one of our participants. In the example in \Cref{tab:qualitative_human}, the LLM incorrectly refers to a black pawn on d5, stating, ``grabbing the d-pawn with 1.Qxd5,'' when the piece on d5 is in fact a black knight. Such hallucinations can confuse the solving player and may lead him to choose an incorrect move.

\section{Human Evaluation Design}
\label{app:human_eval}

\begin{figure}[t]
    \centering
    \includegraphics[width=\columnwidth]{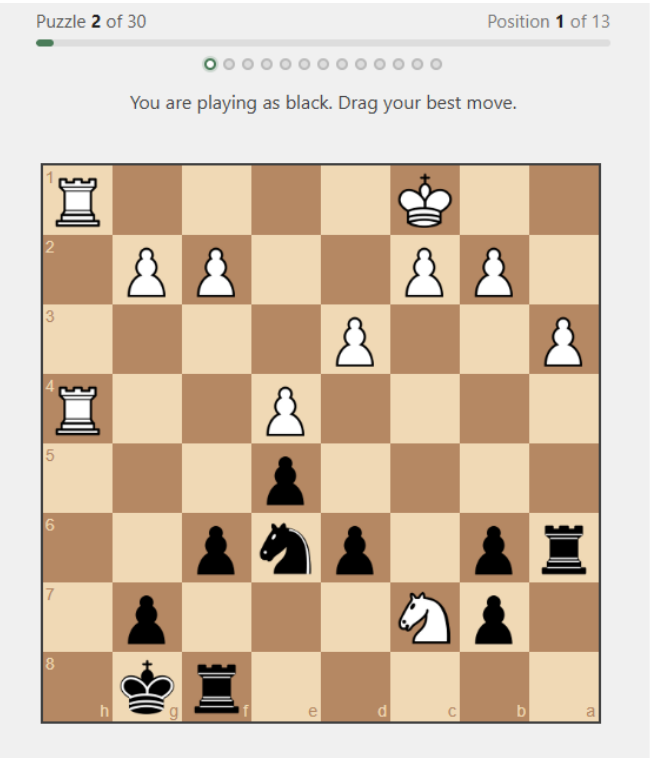}
    \caption{\textbf{Human evaluation user interface for playing puzzles in the No Strategy setting.} Participants submit one move in each board state by dragging the move on the interactive chess board.}
    \label{fig:human_no_strat}
\end{figure}

\begin{figure}[t]
    \centering
    \includegraphics[width=\columnwidth]{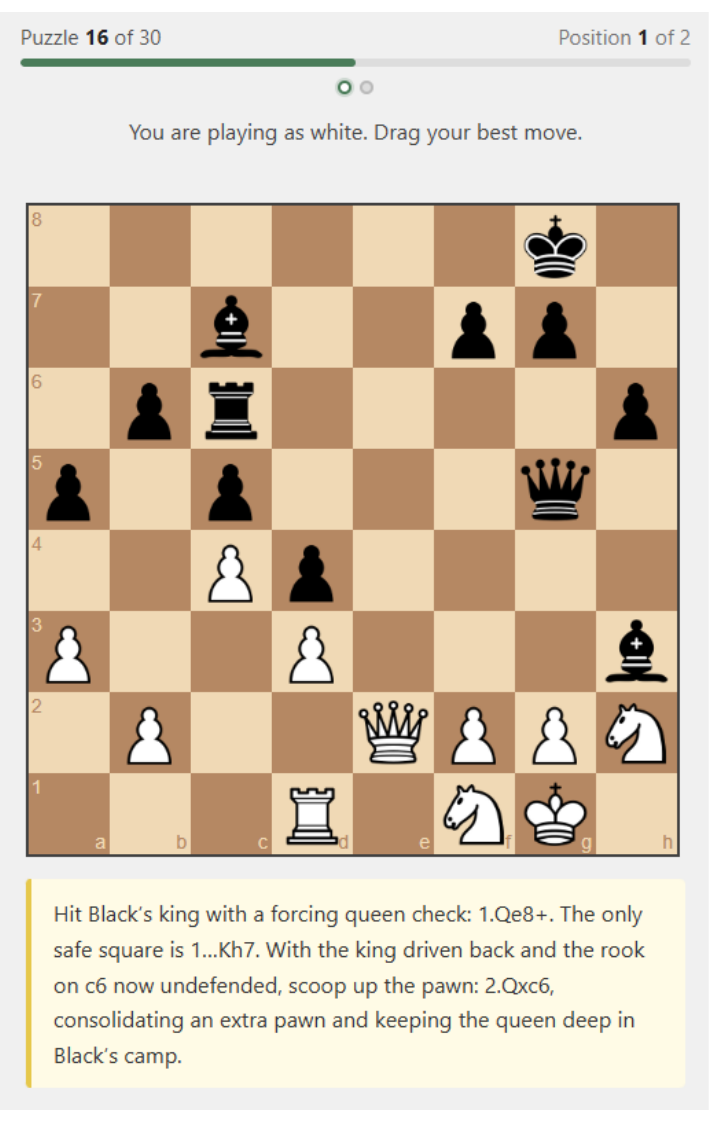}
    \caption{\textbf{Human evaluation user interface for playing puzzles in the Human/LLM setting.} The strategy description is displayed below the corresponding chess board throughout all board states in the puzzle.}
    \label{fig:human_strat}
\end{figure}

\paragraph{Puzzle Configurations.} The participants are asked to play a set of 30 puzzles obtained from \citet{chandler2003chess}. We construct three puzzle configurations, denoted as A, B, and C. Each configuration contains 10 puzzles under each of three settings: no strategy guidance (No Strategy), guidance from a human-written solution in \citet{chandler2003chess} (Human), and guidance from an LLM-generated strategy description in the V-S setting, with $k_{strat}=3$ (LLM). The three different strategies are assigned in a fully balanced and round-robin manner across both configurations and puzzle groups, as follows. In configuration A, puzzles 1--10 are assigned to No Strategy, puzzles 11--20 to Human, and puzzles 21--30 to LLM. In configuration B, puzzles 11--20 are assigned to No Strategy, puzzles 21--30 to Human, and puzzles 1--10 to LLM. In configuration C, puzzles 21--30 are assigned to No Strategy, puzzles 1--10 to Human, and puzzles 11--20 to LLM. Each participant is assigned to one of the configurations A, B, or C. As an individual player's skill affects the evaluation outcome, we recruit participants from a wide range of skill levels and balance the assignment such that each configuration has the same even distribution of players across the three Elo ratings (low, medium, high).

\paragraph{Tree-Expanded Puzzle Playing with Human Players} Our goal is to replace the LLM proxy used as the solving player with an actual human player. To this end, we adapt the Tree-Expanded Puzzle Playing procedure in \Cref{fig:eval_tree} to the human-evaluation setting. For each puzzle, we construct a pruned partial strategy tree with $k_{\mathrm{strat}}=3$, as illustrated on the right of \Cref{fig:strat_tree}. The player nodes in this tree correspond to the board states that can be reached under Tree-Expanded Puzzle Playing with $k_{\mathrm{eval}}\leq 3$, assuming the player plays the optimal move at each player node. For each puzzle, participants are asked to play through all board states corresponding to player nodes in this subtree, by submitting the move they consider best at each state through the interface shown in \Cref{fig:human_no_strat,fig:human_strat}. During analysis, we follow the standard Tree-Expanded Puzzle Playing procedure, with one exception. In step 4, if the player’s move along the branch selected for expansion (i.e., the branch worst for the solving player) does not match the optimal move, we terminate the evaluation early and use the resulting board state as the final state for scoring. This is necessary because, unlike in automatic evaluation, we do not have real-time access to the human player during analysis and therefore cannot query the additional continuations needed to expand the tree further. The resulting score is an upper bound on the player’s performance under full Tree-Expanded Puzzle Playing, since further expansion can only expose additional weaknesses in the player's strategy. Since this differs slightly from the automatic evaluation procedure, we apply the same modified procedure to the LLM player for a fair comparison.

\paragraph{Additional Details}
We recruited a total of 30 university students. During recruitment, participants were informed that their Elo ratings and study responses would be used for research purposes, and informed consent was obtained prior to participation. We do not report any information that uniquely identifies any participants. Each participant was paid the equivalent of around 30 USD for participating in the study, which lasted up to 2 hours per participant. The study protocol was approved by our Institutional Review Board. The instructions provided to the participants are included in the supplementary material. 

\section{LLM Prompts}
\label{app:prompts}

We present representative prompts used for both the verbalization pipeline and the solving player in \Cref{tab:verbalize_prompts,tab:play_prompts,tab:critic_prompts}. The strategy descriptions are generated at the start of each puzzle, and provided throughout puzzle playing. In these prompts, moves are written in algebraic notation following the convention of the Portable Game Notation (PGN) rather than in Universal Chess Interface (UCI) coordinate notation. Algebraic notation is standard in human chess writing, whereas UCI notation is primarily used for communication with chess engines and other chess software. In our experiments, LLMs generated fewer illegal moves when using PGN-style notation, likely because chess-related text in their training data more often appears in algebraic notation than in UCI notation. For ease of parsing, we also instruct the LLMs to return outputs in JSON format; we omit these formatting instructions here for brevity. The exact prompts are provided in the accompanying material released.

\begin{table*}[t]
\centering
\begin{tabular}{p{15cm}}
\hline
\textbf{V-S} \\
\hline
You are analyzing a game of Chess right now. You are given the current state of the board in FEN notation, the current player's color, and a strategy describing the best moves for the current player as well as possible opponent response variations as calculated by a chess engine. The strategy is represented as a JSON tree, and the moves in the strategy are in the Algebraic format. You should provide a high level description of this strategy tree from the perspective of the current player so that another chess player will be able to follow this description and play the strategy as shown here. Keep it short and sweet.\\
FEN:\\
\textbf{[board state in FEN]}\\
Current Player:\\
\textbf{[`Black'/`White']}\\
Strategy:\\
\textbf{[JSON strategy]}\\
\hline
\textbf{V-C}\\
\hline
You are analyzing a game of Chess right now. You are given the current state of the board in FEN notation, the current player's color, and a list of high level chess concepts describing the idea behind the correct strategy. You should provide a coherent description of these concepts from the perspective of the current player so that another chess player will be able to follow this description and play the strategy. Keep it short and sweet.\\
FEN:\\
\textbf{[board state in FEN]}\\
Current Player:\\
\textbf{[`Black'/`White']}\\
Concepts:\\
\textbf{[concept list]}\\
\hline
\textbf{RV-S} \\
\hline
You are analyzing a game of Chess right now. You are given the current state of the board in FEN notation, the current player's color, and a strategy describing the best moves for the current player as well as possible opponent response variations as calculated by a chess engine. The strategy is represented as a JSON tree, and the moves in the strategy are in the Algebraic format. You are also given a previous attempt at describing this strategy in natural language, along with specific feedback from a chess player explaining what was unclear or missing.\\
Rewrite the description so that it addresses every point in the feedback. The result should be a clear, concise strategy description that a chess player can follow without needing to calculate for themselves.\\
FEN:\\
\textbf{[board state in FEN]}\\
Current Player:\\
\textbf{[`Black'/`White']}\\
Strategy:\\
\textbf{[JSON strategy]}\\
Previous Description:\\
\textbf{[generated description from previous round]}\\
Feedback:\\
\textbf{[feedback]}\\
\end{tabular}
\caption{\textbf{Prompts used to verbalize strategies in various settings.} Prompts for the \textbf{V-S+C} setting can be obtained by modifying \textbf{V-S} to include the concept list. Prompts for \textbf{RV-C/RV-S+C} can be obtained by adding the feedback component to \textbf{V-C/V-S+C}, similar to adding the feedback component to \textbf{V-S} to obtain \textbf{RV-S}.}
\label{tab:verbalize_prompts}
\end{table*}

\begin{table*}[t]
\centering
\begin{tabular}{p{15cm}}
\hline
\textbf{No Strategy} \\
\hline
We are playing a game of Chess right now. You are given the current state of the board in FEN notation. You should come up with the next best move in the Algebraic format for the current player. Think about all possible moves you can make, as well as all possible opponent responses before deciding on the best move.\\
FEN:\\
\textbf{[board state in FEN]}\\
Current Player:\\
\textbf{[`Black'/`White']}\\
\hline
\textbf{R} \\
\hline
We are playing a game of Chess right now. You are given the current state of the board in FEN notation. You previously proposed the move \textbf{[previous move]}, but received feedback that it can be improved. Take this feedback into account and provide a better move. \\
FEN: \\
\textbf{[board state in FEN]} \\
Current Player: \\
\textbf{[`Black'/`White']} \\
Previous Move: \\
\textbf{[previous move]} \\
Feedback: \\
\textbf{[feedback]} \\
\hline
\textbf{S or C or (R)V-S/C/S+C} \\
\hline
We are playing a game of Chess right now. You are given the current state of the board in FEN notation. You are also given a description of the strategy by a chess expert, which you should try to follow. You should come up with the next best move in the Algebraic format for the current player.\\
FEN:\\
\textbf{[board state in FEN]}\\
Current Player:\\
\textbf{[`Black'/`White']}\\
Strategy:\\
\textbf{[generated strategy description/JSON strategy tree/concept list]}\\
Previous Sequence of Moves:\\
\textbf{[sequence of moves since the start of the puzzle]}\\
\end{tabular}
\caption{\textbf{Prompts used to sample moves from the LLM player in various settings.}}
\label{tab:play_prompts}
\end{table*}

\begin{table*}[t]
\centering
\begin{tabular}{p{15cm}}
\hline
\textbf{R} \\
\hline
You are a chess expert analyzing a game. You are given the current state of the board in FEN notation, the current player's color, and a move proposed by a chess player.\\
Evaluate whether this is the best move available. Consider:\\
1. Is there a stronger move that creates more threats, wins material, or improves the position?\\
2. Does this move miss any immediate tactics or overlook a better continuation?\\
Provide specific, actionable feedback so the player can find a better move if needed. Decide whether this move is already the best available.\\
FEN:\\
\textbf{[board state in FEN]}\\
Current Player:\\
\textbf{[`Black'/`White']}\\
Proposed Move:\\
\textbf{[move]}\\
\hline
\textbf{RV-S} \\
\hline
You are a chess player who is currently solving a chess puzzle. You are given the starting position of the puzzle in FEN notation, your color, and a natural language description of a strategy written to guide you through this puzzle.\\
Your task is to critically evaluate how useful this strategy description is to a chess player trying to solve the puzzle. Work through the following steps:\\
1. Mentally simulate the main scenarios you would face in this puzzle and try to apply the strategy description to each one.\\
2. Assess how well the description covers those scenarios: is it clear what move to make, does it handle the key opponent responses, and does it avoid ambiguity or missing cases?\\
3. Provide specific, actionable feedback — concrete suggestions for what to add, clarify, or remove so the description becomes more useful. Be precise: if a move or branch is missing, say which one.\\
4. Decide whether the description is already sufficient for a competent chess player to follow the correct strategy without needing to calculate for themselves.\\
You are also provided the full engine strategy tree in JSON format so you can verify whether the description covers all the key variations.\\
FEN:\\
\textbf{[board state in FEN]}\\
Current Player:\\
\textbf{[`Black'/`White']}\\
Strategy Tree:\\
\textbf{[JSON strategy]}\\
Strategy Description:
\textbf{[description]}
\end{tabular}
\caption{\textbf{Prompts used for the LLM critic in various settings.} Prompts for \textbf{RV-C/RV-S+C} can be obtained by replacing the strategy tree with a concept list or adding a concept list to the strategy tree.}
\label{tab:critic_prompts}
\end{table*}

\section{Datasets}
\label{app:dataset}
We include all the datasets used in our evaluation in the supplementary material. These datasets may be used for research purposes only.

\paragraph{Lichess Puzzle Dataset} The Lichess Chess Puzzle Dataset is a set of over 5 million interesting board states ranging from openings to endgames extracted by analyzing 300 million actual games in the Lichess games database using Stockfish 12/13/14/15 NNUE at 40 meganodes \cite{lichessPuzzles}. The resulting puzzles were then automatically tagged with a list of themes, which are further refined by player votes. As the dataset is updated monthly, we include the subset we used in our evaluation in the accompanying material for reproducibility. The full dataset can be downloaded at https://database.lichess.org/\#puzzles. Note that the FEN included in the datasets actually represents the board state one move \textit{before} the actual start of the puzzle. The start of the puzzle can be recovered by playing the first move of the "Moves" column in the dataset.

\paragraph{Dataset for Human Evaluation} We extract 30 puzzles from the book \textit{Chess Tactics for Kids} by \citet{chandler2003chess}. We additionally use the solutions included in the book as the human-written descriptions.

\section{Artifact Usage and Licensing.}
We used Stockfish and Fairy-Stockfish as engines for analysis and move evaluation. Both engines are distributed under the GNU General Public License v3, which permits running the software for research purposes; we do not modify or redistribute the engines as part of our work. The Lichess puzzle dataset is released under the Creative Commons CC0 license, which permits use, modification, redistribution, and publication without additional permission. For puzzles drawn from \textit{Chess Tactics for Kids} \citep{chandler2003chess}, we used the book as a cited research source and did not redistribute the book or its content.

\section{Experimental Hardware}
We ran the Stockfish 16 Chess Engine on an AMD Ryzen 5 5500 CPU with 32 GB RAM. The local instance of \texttt{gpt-oss-120b} was hosted using the vLLM \cite{kwon2023efficient} library on 4 Nvidia H100 GPUs with 80 GB memory each.

%% file: acl_latex.bib
@article{silver2017learning,
    author = {David Silver  and Thomas Hubert  and Julian Schrittwieser  and Ioannis Antonoglou  and Matthew Lai  and Arthur Guez  and Marc Lanctot  and Laurent Sifre  and Dharshan Kumaran  and Thore Graepel  and Timothy Lillicrap  and Karen Simonyan  and Demis Hassabis },
    title = {A general reinforcement learning algorithm that masters chess, shogi, and {Go} through self-play},
    journal = {Science},
    volume = {362},
    pages = {1140-1144},
    year = {2018},
}

@inproceedings{kim2025bridging,
  title={Bridging the gap between expert and language models: Concept-guided chess commentary generation and evaluation},
  author={Kim, Jaechang and Goh, Jinmin and Hwang, Inseok and Cho, Jaewoong and Ok, Jungseul},
  booktitle={Proceedings of the 2025 Conference of the Nations of the Americas Chapter of the Association for Computational Linguistics: Human Language Technologies},
  pages={9497--9516},
  year={2025}
}

@inproceedings{jhamtani2018learning,
    title = "Learning to Generate Move-by-Move Commentary for Chess Games from Large-Scale Social Forum Data",
    author = "Jhamtani, Harsh  and
      Gangal, Varun  and
      Hovy, Eduard  and
      Neubig, Graham  and
      Berg-Kirkpatrick, Taylor",
    booktitle = "Proceedings of the 56th Annual Meeting of the Association for Computational Linguistics",
    year = "2018",
    pages = "1661--1671",
}

@inproceedings{zang2019automated,
  title={Automated chess commentator powered by neural chess engine},
  author={Zang, Hongyu and Yu, Zhiwei and Wan, Xiaojun},
  booktitle={Proceedings of the 57th Annual Meeting of the Association for Computational Linguistics},
  pages={5952--5961},
  year={2019}
}

@inproceedings{papineni2002bleu,
author = {Papineni, Kishore and Roukos, Salim and Ward, Todd and Zhu, Wei-Jing},
title = {{BLEU}: a method for automatic evaluation of machine translation},
year = {2002},
booktitle = {Proceedings of the 40th Annual Meeting of the Association for Computational Linguistics},
pages = {311–318},
}

@article{feng2023chessgpt,
  title={Chess{GPT}: Bridging policy learning and language modeling},
  author={Feng, Xidong and Luo, Yicheng and Wang, Ziyan and Tang, Hongrui and Yang, Mengyue and Shao, Kun and Mguni, David and Du, Yali and Wang, Jun},
  journal={Advances in Neural Information Processing Systems},
  volume={36},
  pages={7216--7262},
  year={2023}
}

@inproceedings{zhang2025complete,
  title={Complete chess games enable {LLM} become a chess master},
  author={Zhang, Yinqi and Han, Xintian and Li, Haolong and Chen, Kedi and Lin, Shaohui},
  booktitle={Proceedings of the 2025 Conference of the Nations of the Americas Chapter of the Association for Computational Linguistics: Human Language Technologies},
  pages={1--7},
  year={2025}
}

@article{kolasani2025llm,
  title={{LLM CHESS}: Benchmarking Reasoning and Instruction-Following in {LLM}s through Chess},
  author={Kolasani, Sai and Saplin, Maxim and Crispino, Nicholas and Montgomery, Kyle and Davis, Jared Quincy and Zaharia, Matei and Wang, Chi and Wang, Chenguang},
  journal={arXiv preprint arXiv:2512.01992},
  year={2025}
}

@article{schultz2024mastering,
  title={Mastering board games by external and internal planning with language models},
  author={Schultz, John and Adamek, Jakub and Jusup, Matej and Lanctot, Marc and Kaisers, Michael and Perrin, Sarah and Hennes, Daniel and Shar, Jeremy and Lewis, Cannada and Ruoss, Anian and others},
  journal={arXiv preprint arXiv:2412.12119},
  year={2024}
}

@article{lee2022improving,
  title={Improving chess commentaries by combining language models with symbolic reasoning engines},
  author={Lee, Andrew and Wu, David and Dinan, Emily and Lewis, Mike},
  journal={arXiv preprint arXiv:2212.08195},
  year={2022}
}

@book{elo1978rating,
  title={The Rating of Chessplayers: Past and Present},
  author={Elo, A.E.},
  isbn={9780923891275},
  lccn={2010549499},
  year={2008},
  publisher={Ishi Press International}
}

@article{campbell2002deep,
  title={Deep {B}lue},
  author={Campbell, Murray and Hoane Jr, A Joseph and Hsu, Feng-hsiung},
  journal={Artificial intelligence},
  volume={134},
  number={1-2},
  pages={57--83},
  year={2002},
  publisher={Elsevier}
}

@misc{stockfish,
  title        = {Stockfish},
  author       = {Romstad, Tord and Costalba, Marco and Kiiski, Joona and The Stockfish Community},
  url          = {https://stockfishchess.org/},
  note         = {A free and strong UCI chess engine},
  year         = {2024}
}

@inproceedings{luo2024repoagent,
  title={Repo{A}gent: An {LLM}-powered open-source framework for repository-level code documentation generation},
  author={Luo, Qinyu and Ye, Yining and Liang, Shihao and Zhang, Zhong and Qin, Yujia and Lu, Yaxi and Wu, Yesai and Cong, Xin and Lin, Yankai and Zhang, Yingli and others},
  booktitle={Proceedings of the 2024 Conference on Empirical Methods in Natural Language Processing: System Demonstrations},
  pages={436--464},
  year={2024}
}

@inproceedings{yang2025docagent,
  title={Doc{A}gent: A multi-agent system for automated code documentation generation},
  author={Yang, Dayu and Simoulin, Antoine and Qian, Xin and Liu, Xiaoyi and Cao, Yuwei and Teng, Zhaopu and Yang, Grey},
  booktitle={Proceedings of the 63rd Annual Meeting of the Association for Computational Linguistics (Volume 3: System Demonstrations)},
  pages={460--471},
  year={2025}
}

@inproceedings{iyer2016summarizing,
    title = "Summarizing Source Code using a Neural Attention Model",
    author = "Iyer, Srinivasan  and
      Konstas, Ioannis  and
      Cheung, Alvin  and
      Zettlemoyer, Luke",
    booktitle = "Proceedings of the 54th Annual Meeting of the Association for Computational Linguistics",
    year = "2016",
    pages = "2073--2083"
}

@inproceedings{hu2018deepcom,
author = {Hu, Xing and Li, Ge and Xia, Xin and Lo, David and Jin, Zhi},
title = {Deep code comment generation},
year = {2018},
booktitle = {Proceedings of the 26th Conference on Program Comprehension},
pages = {200–210},
}

@inproceedings{feng2020codebert,
  title={Code{BERT}: A pre-trained model for programming and natural languages},
  author={Feng, Zhangyin and Guo, Daya and Tang, Duyu and Duan, Nan and Feng, Xiaocheng and Gong, Ming and Shou, Linjun and Qin, Bing and Liu, Ting and Jiang, Daxin and others},
  booktitle={Findings of the Association for Computational Linguistics: EMNLP 2020},
  pages={1536--1547},
  year={2020}
}

@article{madaan2023self,
  title={Self-{R}efine: Iterative refinement with self-feedback},
  author={Madaan, Aman and Tandon, Niket and Gupta, Prakhar and Hallinan, Skyler and Gao, Luyu and Wiegreffe, Sarah and Alon, Uri and Dziri, Nouha and Prabhumoye, Shrimai and Yang, Yiming and others},
  journal={Advances in Neural Information Processing Systems},
  volume={36},
  pages={46534--46594},
  year={2023}
}

@inproceedings{liu2023g,
  title={G-{E}val: {NLG} evaluation using {GPT}-4 with better human alignment},
  author={Liu, Yang and Iter, Dan and Xu, Yichong and Wang, Shuohang and Xu, Ruochen and Zhu, Chenguang},
  booktitle={Proceedings of the 2023 Conference on Empirical Methods in Natural Language Processing},
  pages={2511--2522},
  year={2023}
}

@misc{openai_o3_2025,
  author       = {{OpenAI}},
  title        = {{OpenAI o3}},
  url          = {https://openai.com/index/introducing-o3-and-o4-mini/},
  year         = {2025},
  note         = {Model snapshot: o3-2025-04-16}
}

@article{agarwal2025gpt,
  title={gpt-oss-120b \& gpt-oss-20b model card},
  author={Agarwal, Sandhini and Ahmad, Lama and Ai, Jason and Altman, Sam and Applebaum, Andy and Arbus, Edwin and Arora, Rahul K and Bai, Yu and Baker, Bowen and Bao, Haiming and others},
  journal={arXiv preprint arXiv:2508.10925},
  year={2025}
}

@inproceedings{kwon2023efficient,
  title={Efficient Memory Management for Large Language Model Serving with {P}aged{A}ttention},
  author={Woosuk Kwon and Zhuohan Li and Siyuan Zhuang and Ying Sheng and Lianmin Zheng and Cody Hao Yu and Joseph E. Gonzalez and Hao Zhang and Ion Stoica},
  booktitle={Proceedings of the ACM SIGOPS 29th Symposium on Operating Systems Principles},
  year={2023}
}

@misc{arena2025,
  author       = {{Kaggle}},
  title        = {{Kaggle Game Arena}},
  url          = {https://www.kaggle.com/game-arena},
  year         = {2025},
}

@misc{lichessPuzzles,
  author       = {{Lichess}},
  title        = {{Lichess chess puzzle dataset}},
  url          = {https://huggingface.co/datasets/Lichess/chess-puzzles},
  year         = {2026},
}

@misc{lichessAccuracy,
  author       = {{Lichess}},
  title        = {{Lichess Accuracy metric}},
  url          = {https://lichess.org/page/accuracy},
  year         = {2026},
}

@book{chandler2003chess,
  title={Chess Tactics for Kids},
  author={Chandler, M.},
  isbn={9781901983999},
  url={https://books.google.com.sg/books?id=lrsEAAAACAAJ},
  year={2003},
  publisher={Gambit}
}

@inproceedings{xu2025ai,
  title={A{I} self-preferencing in algorithmic hiring: Empirical evidence and insights},
  author={Xu, Jiannan and Li, Gujie and Jiang, Jane Yi},
  booktitle={Proceedings of the AAAI/ACM Conference on AI, Ethics, and Society},
  pages={2757--2758},
  year={2025}
}

@misc{anand2026interview,
  author       = {Wait, Theo and Anand, Viswanathan},
  title        = {Viswanathan {A}nand on His Legacy, {M}agnus, {G}ukesh, {S}indarov and the Future of Chess},
  year         = {2026},
  note         = {Interview. Quote at 24:58},
  url          = {https://www.youtube.com/watch?v=yvsKQa18VwY&t=1498s}
}

@inproceedings{timbers2022approximate,
  title={Approximate Exploitability: Learning a Best Response},
  author={Timbers, Finbarr and Bard, Nolan and Lockhart, Edward and Lanctot, Marc and Schmid, Martin and Burch, Neil and Schrittwieser, Julian and Hubert, Thomas and Bowling, Michael},
  booktitle={Proceedings of the International Joint Conference on
               Artificial Intelligence},
  pages={3487--3493},
  year={2022}
}

@article{lanctot2017unified,
  title={A unified game-theoretic approach to multiagent reinforcement learning},
  author={Lanctot, Marc and Zambaldi, Vinicius and Gruslys, Audrunas and Lazaridou, Angeliki and Tuyls, Karl and P{\'e}rolat, Julien and Silver, David and Graepel, Thore},
  journal={Advances in neural information processing systems},
  volume={30},
  year={2017}
}

@article{hsu1990grandmaster,
  title={A grandmaster chess machine},
  author={Hsu, Feng-hsiung and Anantharaman, Thomas and Campbell, Murray and Nowatzyk, Andreas},
  journal={Scientific American},
  volume={263},
  number={4},
  pages={44--51},
  year={1990},
  publisher={JSTOR}
}

@misc{fairystockfish,
  title        = {Fairy-{S}tockfish},
  author       = {Fichter, Fabian and The Fairy-Stockfish Community},
  url          = {https://fairy-stockfish.github.io/},
  note         = {A chess variant engine derived from Stockfish},
  year         = {2021}
}

@article{chen2025xiangqi,
  title={Xiangqi-r1: Enhancing spatial strategic reasoning in {LLM}s for {Chinese} chess via reinforcement learning},
  author={Chen, Yuhao and Liu, Shuochen and Lyu, Yuanjie and Zhang, Chao and Shi, Jiayao and Xu, Tong},
  journal={arXiv preprint arXiv:2507.12215},
  year={2025}
}

@article{brown2018superhuman,
  title={Superhuman {AI} for heads-up no-limit poker: {L}ibratus beats top professionals},
  author={Brown, Noam and Sandholm, Tuomas},
  journal={Science},
  volume={359},
  number={6374},
  pages={418--424},
  year={2018},
}

@article{brown2019superhuman,
  title={Superhuman {AI} for multiplayer poker},
  author={Brown, Noam and Sandholm, Tuomas},
  journal={Science},
  volume={365},
  number={6456},
  pages={885--890},
  year={2019},
}

@article{meta2022human,
  title={Human-level play in the game of diplomacy by combining language models with strategic reasoning},
  author={Bakhtin, Anton and Brown, Noam and Dinan, Emily and Farina, Gabriele and Flaherty, Colin and Fried, Daniel and Goff, Andrew and Gray, Jonathan and Hu, Hengyuan and others},
  journal={Science},
  volume={378},
  number={6624},
  pages={1067--1074},
  year={2022},
  publisher={American Association for the Advancement of Science}
}

@article{xu2023exploring,
  title={Exploring large language models for communication games: An empirical study on {W}erewolf},
  author={Xu, Yuzhuang and Wang, Shuo and Li, Peng and Luo, Fuwen and Wang, Xiaolong and Liu, Weidong and Liu, Yang},
  journal={arXiv preprint arXiv:2309.04658},
  year={2023}
}

@article{light2023avalonbench,
  title={Avalon{B}ench: Evaluating {LLM}s playing the game of {A}valon},
  author={Light, Jonathan and Cai, Min and Shen, Sheng and Hu, Ziniu},
  journal={arXiv preprint arXiv:2310.05036},
  year={2023}
}

@article{carminati2023hidden,
  title={Hidden-role games: {E}quilibrium concepts and computation},
  author={Carminati, Luca and Zhang, Brian Hu and Farina, Gabriele and Gatti, Nicola and Sandholm, Tuomas},
  journal={arXiv preprint arXiv:2308.16017},
  year={2023}
}

@article{tang2024maia,
  title={Maia-2: A unified model for human-{AI} alignment in chess},
  author={Tang, Zhenwei and Jiao, Difan and McIlroy-Young, Reid and Kleinberg, Jon and Sen, Siddhartha and Anderson, Ashton},
  journal={Advances in Neural Information Processing Systems},
  volume={37},
  pages={20919--20944},
  year={2024}
}

@article{siu2021evaluation,
  title={Evaluation of human-{AI} teams for learned and rule-based agents in hanabi},
  author={Siu, Ho Chit and Pe{\~n}a, Jaime and Chen, Edenna and Zhou, Yutai and Lopez, Victor and Palko, Kyle and Chang, Kimberlee and Allen, Ross},
  journal={Advances in Neural Information Processing Systems},
  volume={34},
  pages={16183--16195},
  year={2021}
}

@inproceedings{mcilroy2020aligning,
  title={Aligning superhuman {AI} with human behavior: Chess as a model system},
  author={McIlroy-Young, Reid and Sen, Siddhartha and Kleinberg, Jon and Anderson, Ashton},
  booktitle={Proceedings of the 26th ACM SIGKDD International Conference on Knowledge Discovery \& Data Mining},
  pages={1677--1687},
  year={2020}
}

@article{shih2021critical,
  title={On the critical role of conventions in adaptive human-{AI} collaboration},
  author={Shih, Andy and Sawhney, Arjun and Kondic, Jovana and Ermon, Stefano and Sadigh, Dorsa},
  journal={arXiv preprint arXiv:2104.02871},
  year={2021}
}

@misc{icga2024wccc50,
  author  = {{International Computer Games Association}},
  title   = {{2024 World Computer Chess Championships: The 50th Anniversary}},
  year    = {2024},
  url     = {https://icga.org/?page_id=3957},
  urldate = {2026-05-25}
}

@article{mcgrath2022acquisition,
  title={Acquisition of chess knowledge in {A}lpha{Z}ero},
  author={McGrath, Thomas and Kapishnikov, Andrei and Toma{\v{s}}ev, Nenad and Pearce, Adam and Wattenberg, Martin and Hassabis, Demis and Kim, Been and Paquet, Ulrich and Kramnik, Vladimir},
  journal={Proceedings of the National Academy of Sciences},
  volume={119},
  number={47},
  pages={e2206625119},
  year={2022},
  publisher={National Academy of Sciences}
}

@article{puri2019explain,
  title={Explain your move: Understanding agent actions using specific and relevant feature attribution},
  author={Puri, Nikaash and Verma, Sukriti and Gupta, Piyush and Kayastha, Dhruv and Deshmukh, Shripad and Krishnamurthy, Balaji and Singh, Sameer},
  journal={arXiv preprint arXiv:1912.12191},
  year={2019}
}

@inproceedings{fritz2021some,
  title={Some chess-specific improvements for perturbation-based saliency maps},
  author={Fritz, Jessica and F{\"u}rnkranz, Johannes},
  booktitle={2021 IEEE Conference on Games (CoG)},
  pages={01--08},
  year={2021},
  organization={IEEE}
}

@article{spinnato2025towards,
  title={Towards Piece-by-Piece Explanations for Chess Positions with {SHAP}},
  author={Spinnato, Francesco},
  journal={arXiv preprint arXiv:2510.25775},
  year={2025}
}

@article{schut2025bridging,
  title={Bridging the human--{AI} knowledge gap through concept discovery and transfer in {A}lpha{Z}ero},
  author={Schut, Lisa and Toma{\v{s}}ev, Nenad and McGrath, Thomas and Hassabis, Demis and Paquet, Ulrich and Kim, Been},
  journal={Proceedings of the National Academy of Sciences},
  volume={122},
  number={13},
  pages={e2406675122},
  year={2025},
  publisher={National Academy of Sciences}
}

@inproceedings{palsson2023unveiling,
  title={Unveiling Concepts Learned by a World-Class Chess-Playing Agent.},
  author={P{\'a}lsson, A{\dh}alsteinn and Bj{\"o}rnsson, Yngvi},
  booktitle={Proceedings of the International Joint Conference on Artificial Intelligence},
  pages={4864--4872},
  year={2023}
}

@article{chen2026extracting,
  title={Extracting Search Trees from {LLM} Reasoning Traces Reveals Myopic Planning},
  author={Chen, Sixing and Li, Ji-An and Cakir, Saner and Akcali, Sinan and Lee, Kayla and Mattar, Marcelo G},
  journal={arXiv preprint arXiv:2605.06840},
  year={2026}
}

@inproceedings{johanson2011accelerating,
  title={Accelerating best response calculation in large extensive games},
  author={Johanson, Michael and Waugh, Kevin and Bowling, Michael and Zinkevich, Martin},
  booktitle={Proceedings of the International Joint Conference on Artificial Intelligence},
  pages={258--265},
  year={2011}
}

@misc{chesscom,
  author       = {{Chess.com}},
  title        = {{Chess.com: Play Chess Online on the \#1 Site}},
  year         = {2026},
  howpublished = {\url{https://www.chess.com/}},
  note         = {Accessed: 2026-05-26}
}
